\documentclass[10pt,twocolumn,letterpaper]{article}
\usepackage{amsthm}

\usepackage{iccv}              

\definecolor{iccvblue}{rgb}{0.21,0.49,0.74}
\usepackage[pagebackref,breaklinks,colorlinks,allcolors=iccvblue]{hyperref}

\def\paperID{11313} 
\def\confName{ICCV}
\def\confYear{2025}

\title{Sparse Fine-Tuning of Transformers for Generative Tasks}

\author{Wei Chen, Jingxi Yu, Zichen Miao, Qiang Qiu \\
Purdue University, IN, USA\\
\texttt{\{chen2732, yu667, miaoz, qqiu\}@purdue.edu}
}

\begin{document}

\newtheorem{theorem}{Theorem}[section]
\newtheorem{proposition}[theorem]{Proposition}
\newtheorem{lemma}[theorem]{Lemma}
\newtheorem{corollary}[theorem]{Corollary}
\newtheorem{definition}[theorem]{Definition}
\newtheorem{assumption}[theorem]{Assumption}
\newtheorem{remark}[theorem]{Remark}

\newcommand{\Wq}{\mathbf{W}_q}
\newcommand{\Wk}{\mathbf{W}_k}
\newcommand{\Wv}{\mathbf{W}_v}
\newcommand{\Wo}{\mathbf{W}_o}
\newcommand{\Wvo}{\mathbf{W}_{vo}}
\newcommand{\Q}{\mathbf{Q}}
\newcommand{\K}{\mathbf{K}}
\newcommand{\V}{\mathbf{V}}
\newcommand{\W}{\mathbf{W}}
\newcommand{\attn}{\mathbf{A}}
\newcommand{\inp}{\mathbf{X}}
\newcommand{\inpv}{\mathbf{x}}
\newcommand{\out}{\mathbf{O}}
\newcommand{\feat}{\mathbf{Z}}
\newcommand{\filter}{\mathbf{F}}
\newcommand{\coeff}{\mathbf{W}_s}
\newcommand{\atoms}{\mathbf{D}}

\maketitle
\begin{abstract}
Large pre-trained transformers have revolutionized artificial intelligence across various domains, and fine-tuning remains the dominant approach for adapting these models to downstream tasks due to the cost of training from scratch. 
However, in existing fine-tuning methods, the updated representations are formed as a dense combination of modified parameters, making it challenging to interpret their contributions and understand how the model adapts to new tasks.
In this work, we introduce a fine-tuning framework inspired by sparse coding, where fine-tuned features are represented as a sparse combination of basic elements, i.e., feature dictionary atoms. 
The feature dictionary atoms function as fundamental building blocks of the representation, and tuning atoms allows for seamless adaptation to downstream tasks.
Sparse coefficients then serve as indicators of atom importance, identifying the contribution of each atom to the updated representation.
Leveraging the atom selection capability of sparse coefficients, we first demonstrate that our method enhances image editing performance by improving text alignment through the removal of unimportant feature dictionary atoms.
Additionally, we validate the effectiveness of our approach in the text-to-image concept customization task, where our method efficiently constructs the target concept using a sparse combination of feature dictionary atoms, outperforming various baseline fine-tuning methods.

%
%
\end{abstract}    
\section{Introduction}
\label{sec:intro}

 


In recent years, pre-trained large models have revolutionized artificial intelligence across domains like natural language processing and computer vision~\cite{vit, vaswani2017attention, rombach2022high, sam}. Given the prohibitive computational resources required for training these models from scratch, fine-tuning has emerged as the dominant paradigm for adapting pre-trained models to specific downstream tasks, enabling efficient knowledge transfer while minimizing computational overhead.

Various parameter-efficient fine-tuning (PEFT) methods~\cite{hu2021lora, liu2024dora, oft} have been developed to tune the large transformers by updating a small subset of parameters while keeping large pre-trained weights frozen. 
A representative family of these approaches~\cite{hu2021lora, liu2024dora} utilizes lightweight parameterizations, such as low-rank decomposition, to adapt the weight matrices.
However, despite being lightweight, these methods obtain the updated features with a dense matrix multiplication between the input and the adapted weights, making it challenging to interpret how the individual updated parameters contribute to the newly learned representations and to identify the new knowledge gained through fine-tuning.

To tackle this challenge, we take inspiration from sparse coding~\cite{mairal2008supervised, olshausen1996emergence, qiu2018dcfnet, wang2021adaptive, clatoms, l3net, miao2021spatiotemporal, wang2019stochastic, wang2021image, patel2024efficient}, in which signals are expressed as a sparse combination of basic elements, \textit{i.e.}, \textit{dictionary atoms}. 
In this paper, we frame model fine-tuning as a sparse coding problem, where the adapted feature is constructed using a dictionary of feature atoms.
These feature atoms are selectively combined with \textit{sparse coefficients}, ensuring that only a small subset of atoms contributes to the updated representation. This sparsity not only disentangles interactions between features but also allows each atom to capture a distinct aspect of the downstream data.



We first analyze with a toy experiment that the tuned feature dictionary atoms function as building blocks of the adapted representation, and tuning these atoms allows for effective adaptation to downstream tasks. 
We demonstrate that with pre-trained coefficients and atoms, fine-tuning the atoms effectively captures the representation of downstream tasks. Representing features through individual atoms enhances controllability in the adaptation process.

\begin{figure*}[t]
  \centering
 \begin{subfigure}[b]{0.34\textwidth}
     \centering
     \includegraphics[trim={0pt 0pt 540pt 140pt}, clip, width=\textwidth]{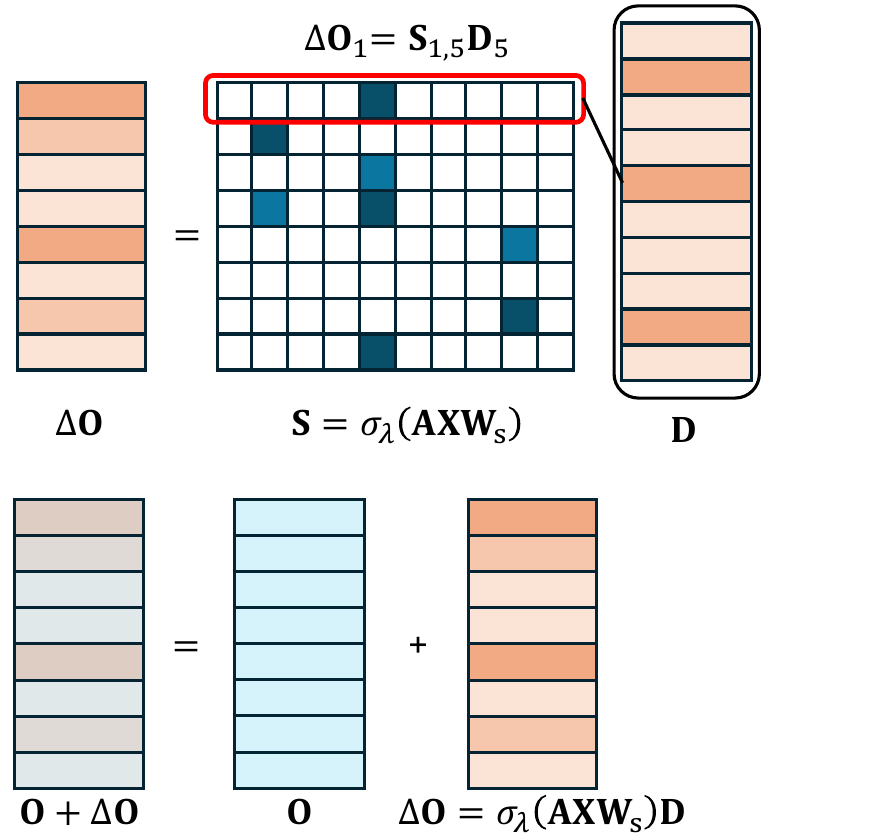}
    \caption{}
 \end{subfigure}
 \begin{subfigure}[b]{0.34\textwidth}
     \centering
     \includegraphics[trim={0pt 0pt 550pt 140pt}, clip, width=\textwidth]{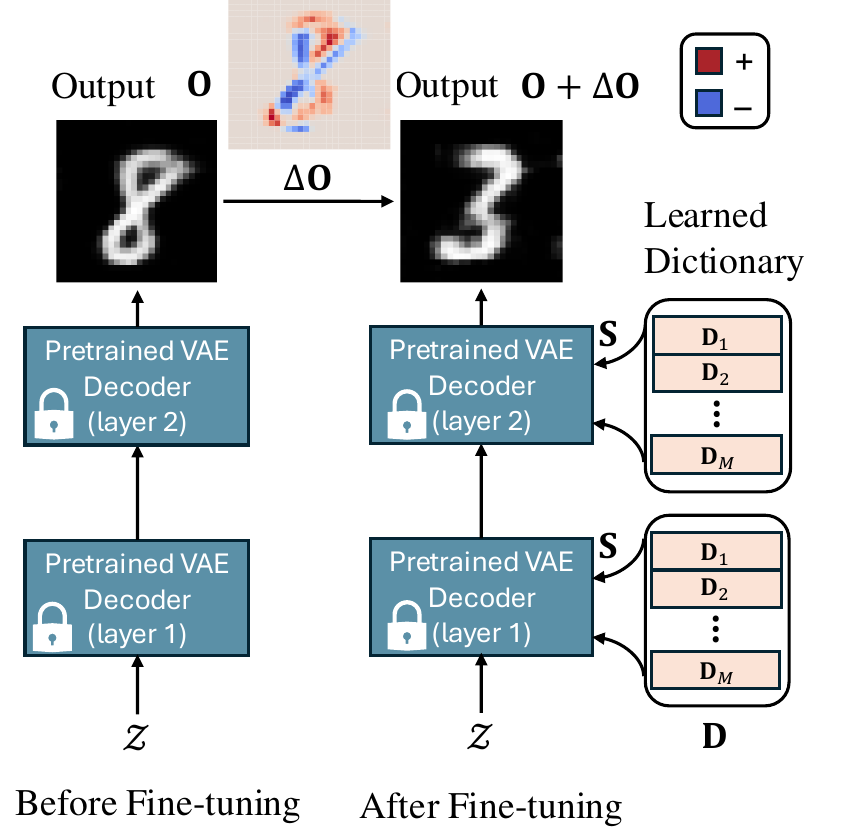}
    \caption{}
 \end{subfigure}
 \begin{subfigure}[b]{0.3\textwidth}
     \centering
     \includegraphics[trim={0pt 0pt 640pt 180pt}, clip, width=\textwidth]{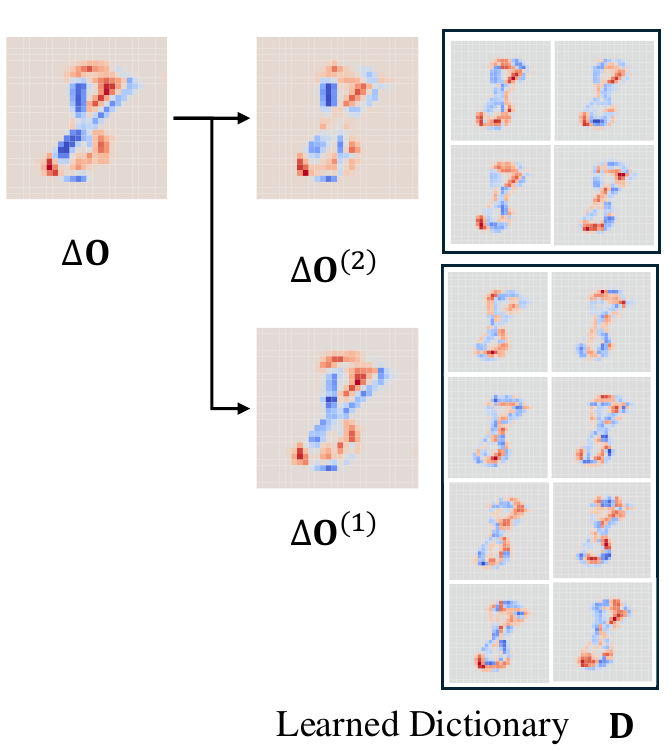}
    \caption{}
 \end{subfigure}
 
  \caption{(a) Our method leverages sparse coding and enforces high sparsity in adapted feature $\Delta \out$, ensuring that the output feature is constructed using only a few dictionary atoms $\atoms$.
  (b) The pre-trained model initially generates ``8". It generates ``3" after integrating the learned dictionary $\atoms$ which is fine-tuned on ``3". 
  (c) The overall transformation of the generated digit from ``8" to ``3" using the dictionary atoms. The red indicates an increase in value at a specific location and blue indicates a decrease.
  The total influence is decomposed into contributions from atoms at different layers: $\Delta \out^{(1)}$ represents the combined effect of 8 atoms in the first layer, while $\Delta \out^{(2)}$ accounts for the 4 atoms in the second layer. The details of this experiment is provided in Section~\ref{sec:alyz_atoms}
  }
  \label{fig:method_ft}
  \vspace{-8pt}
\end{figure*}

We validate our method on generative fine-tuning tasks. We show that the target concept can be effectively represented using feature dictionary atoms after fine-tuning the pre-trained model within our framework. 
Compared to baseline methods, our approach allows the selection of specific atoms to refine the generated image. This adaptability improves performance on image editing tasks by enhancing editing results through the removal of unimportant atoms. 
We also apply our approach to customized text-to-image tasks, showing that fine-tuning atoms effectively captures personal concepts while preserving the capabilities of pre-trained models. Our method outperforms various baseline approaches, demonstrating its effectiveness in controlled 
adaptation for generative modeling.

We summarize our contributions as follows:
\begin{itemize}
    \item We formulate large transformer fine-tuning as a sparse coding problem, where the adapted feature representation is a sparse combination of a dictionary of feature atoms.
    \item We introduce a fine-tuning framework that leverages a combination of dictionary atoms and demonstrates that tuning atoms enhances control over the adaptation process.
    \item We validate the effectiveness of our approach in generative tasks by demonstrating the improvement in image editing and concept customization.
\end{itemize}

\section{Related Work}
\vspace{-3pt}
\label{sec:related_work}

\paragraph{Sparse Representation and Attention}
Recent research has demonstrated the significant role of sparsity in enhancing both interpretability and computational efficiency in generative models through various approaches including sparse autoencoders for feature extraction~\cite{surkov2024unpacking}, sparse feature circuits for improved generalization~\cite{marks2024sparse}, transcoders for detailed circuit analysis~\cite{dunefsky2024transcoders}, dictionary learning for monosemantic representations~\cite{bricken2023monosemanticity}, and selective parameter tuning strategies~\cite{hu2024sara}. Complementary work has addressed the quadratic computational complexity of standard attention mechanisms through sparse attention techniques that compute attention scores for only a subset of token pairs~\cite{child2019generating, vaswani2017attention, beltagy2020longformer, tay2020sparse}, with optimizations including dynamic pattern adjustment~\cite{lai2025flexprefill}, hardware-aligned implementations~\cite{2502.11089}, and graph-based conceptualizations~\cite{tomczak2025longer}, collectively enabling transformers to process significantly longer sequences while maintaining model quality despite challenges in balancing efficiency with performance in complex long-range dependency scenarios. 
Compared to previous works, our method applies sparse coding for fine-tuning transformer-based models in generative tasks.

\vspace{-6pt}
\paragraph{Representation Tuning}
Recent representation tuning approaches operate directly on model representations rather than weights, achieving parameter efficiency while enhancing controllability, and interpretability across various model types and tasks~\cite{chen2024large, miao2025coeff, wu2025reft, wu2024advancing, liu2025re, balasubramanian2024decomposing}. Unlike traditional PEFT approaches that impact representations indirectly through weight adjustments, representation tuning methods preserve the original pre-trained model weights while directly modifying the output representations. 
In comparison, our work studies the sparsity in the updated representation and frames it as a dictionary learning problem.

\vspace{-6pt}
\paragraph{Personalized Text-to-Image Generation}
Extensive research has been conducted on aligning text prompts and generated concepts~\cite{kumari2023multi, ruiz2023dreambooth, miao2024tuning, miao2024training, kim2024learning, purohit2024posterior, kim2025text}. \citet{zhang2024object} developed an object-conditioned energy-based attention map alignment technique in diffusion models, enhancing control over specific objects in generated images. Another study by \citet{zhang2024enhancing} improved semantic fidelity in text-to-image synthesis through attention regulation. \citet{hao2024conceptexpress} introduces ConceptExpress, an unsupervised method for extracting personalized concepts from a single image. \citet{ren2024unveiling} addressed memorization issues in text-to-image diffusion models by analyzing cross-attention mechanisms, proposing solutions to reduce overfitting to user-specific data.
We approach this problem by adopting the ideas from sparse coding and learning the target concept as a sparse combination of coefficients and dictionary atoms.

\section{Method}

\subsection{Preliminary}
\paragraph{Multi-head attention (MHA).}
The attention layer~\cite{vaswani2017attention} transforms the input sequence $\inp \in\mathbb{R}^{N\times C_i}$ to the output sequence $\out \in\mathbb{R}^{N\times C_o}$, where $N$ denotes the sequence length, $C_i$ and $C_o$ are the dimension of input and output features. 
The attention layer projects the input as $\Q = \inp \Wq$, $\K = \inp \Wk$, $\V = \inp \Wv$ using the corresponding projection matrices $\Wq, \Wk, \Wv \in \mathbb{R}^{C_i\times C_o}$, and calculates the attention map,
\begin{equation}
\begin{split}
    \attn = \text{attn}(\inp) &=\text{softmax}(\inp \Wq \Wk^{\intercal} \inp^{\intercal}).
\end{split}
\label{eq:attn_map}
\end{equation}
The attention map $\attn \in \mathbb{R}^{N \times N}$ captures the token-wise relationship by doing inner-product in a space transformed by $\Wq, \Wk$.
Then the output of the attention layer is,
\begin{equation}
\begin{split}
    \out & = \attn \inp \Wv.
\end{split}
\label{eq:attn}
\end{equation}
Multi-head attention extends this by allowing multiple attention mechanisms to work in parallel, with each head independently learning attention patterns. 
The project matrices for each head are $\Wq^{(h)}, \Wk^{(h)}, \Wv^{(h)} \in \mathbb{R}^{C_i\times \frac{C_o}{H}}$, where $h=1,\cdots, H$ and $H$ is the number of heads. 
The multi-head attention represents as,
\begin{equation}
\begin{split}
    \feat & = \sum_{h=1}^{H} \attn^{(h)} \inp \Wvo^{(h)},
\label{eq:multi_head_attn}
\end{split}
\end{equation}
where $\Wo \in \mathbb{R}^{C_o\times C_o}$ and $\Wvo^{(h)}=\Wv^{(h)} \Wo^{(h)\intercal}$, $\Wo^{(h)} \in \mathbb{R}^{C_o\times \frac{C_o}{H} }$, $\Wvo^{(h)} \in \mathbb{R}^{C_i \times C_o}$.

\paragraph{Sparse coding.}
Sparse coding is a signal representation technique that models a signal as a linear combination of a small number of basis functions from a dictionary. Given a signal $\mathbf{X} \in \mathbb{R}^{N \times C_i}$, the goal of sparse coding is to find a dictionary $\mathbf{D} \in \mathbb{R}^{m \times C_i}$ (where $m > C_i$ is the number of atoms) and a sparse coefficient vector $\mathbf{S} \in \mathbb{R}^{N \times m}$ such that the signal can be approximated as:
\begin{equation}
    \mathbf{X} \approx \mathbf{S} \mathbf{D},
\end{equation}
where $\mathbf{S}$ has only a few nonzero elements, ensuring sparsity. The sparse coding problem is often formulated as an optimization problem:
\begin{equation}
    \min_{\mathbf{S}} \|\mathbf{X} - \mathbf{S} \mathbf{D} \|_2^2 + \lambda \|\mathbf{S}\|_1,
\end{equation}
where $\lambda$ is a regularization parameter controlling the trade-off between sparsity and reconstruction error. 
The Proximal gradient method is a common approach to solving this optimization problem, including Iterative Shrinkage-Thresholding Algorithm (ISTA) and its accelerated version (FISTA)~\citep{beck2009fast}, which utilize proximal operators for efficient iterative updates. 
For sparse coding with $\ell_1$ regularization, the corresponding proximal operator is the soft-thresholding function:
\begin{equation}
    \text{prox}_{\lambda \|\cdot\|_1} (\mathbf{S}) = \text{sign}(\mathbf{S}) \odot \max(|\mathbf{S}| - \lambda, 0),
\end{equation}
where $\odot$ is Hadamard product. It encourages sparsity by shrinking small values of $\mathbf{S}$ toward zero. 
Sparse coding is widely used in signal processing, machine learning, and neuroscience, providing efficient and interpretable representations of data.

\subsection{Sparse Fine-Tuning}
In this work, we conceptualize model fine-tuning as a sparse coding problem, i.e., the adapted features are constructed using a dictionary of feature atoms. 
%
As our method focuses on the transformer architecture, we first propose an interpretation of the attention operation by viewing the product between the attention map and value matrices as a combination of dictionary atoms with dictionary coefficients. We then propose a fine-tuning approach by representing the adapted features as a learned sparse combination of dictionary atoms with sparse coefficients, representing the fine-tuned representations' underlying sparsity nature.
%



\paragraph{A dictionary view of attention.} 
For single head attention (\ref{eq:attn}), each row of $\attn$ can be interpreted as coefficients for combining $\mathbf{V}$ into the output $\mathbf{O}$, \textit{i.e.}, $\mathbf{O}_i = \sum_j \attn_{ij} \mathbf{V}_j$.
In the case of multi-head attention (\ref{eq:multi_head_attn}),
\begin{equation}
\begin{split}
    \sum_{h=1}^{H} \attn^{(h)} \inp \Wvo^{(h)} 
    &= 
    \begin{bmatrix}
    \attn^{(1)} & \cdots & \attn^{(H)}
    \end{bmatrix}
    \begin{bmatrix}
    \inp \Wvo^{(1)} \\ 
    \vdots \\ 
    \inp \Wvo^{(H)}
    \end{bmatrix},
\end{split}
\end{equation}
it can be represented as a product of a composite coefficient $\begin{bmatrix}\attn^{(1)}, \cdots, \attn^{(H)}\end{bmatrix} \in \mathbb{R}^{N \times NH}$, and an extensive dictionary $\begin{bmatrix}\inp \Wvo^{(1)}, \cdots, \inp \Wvo^{(H)}\end{bmatrix}^{\intercal} \in \mathbb{R}^{NH \times C_o}$. 
Each row of the output $\mathbf{O}$ is a linear combination of $\inp \Wvo$, \textit{i.e.}, $\mathbf{O}_i = \sum_{h, j} \attn_{ij}^{(h)} (\inp \Wvo^{(h)})_{j}$.
When represented as a linear combination of dictionary elements, multi-head attention and single-head attention follow the same formulation. In the following discussion, we focus on the single-head attention formulation, which can be seamlessly extended to the multi-head attention case.

\paragraph{A sparse dictionary view of attention.}


From the perspective of sparse coding, the feature is expressed as $\mathbf{S} \atoms$, where $\mathbf{S}$ represents the \textit{sparse} coefficients and $\atoms$ consists of dictionary atoms that remain \textit{data-independent} once learned. 
Although attention can be expressed as a linear combination of atoms, it does not inherently exhibit sparsity.
To fully leverage the advantages of sparse representations, we propose to represent the attention mechanism (\ref{eq:attn}) in the form of sparse coding,
\begin{equation}
\begin{split}
    \out = \mathbf{S} \atoms 
    =\sigma_{\lambda}(\attn \inp \coeff) \atoms,
\end{split}
\end{equation}
where $\sigma_{\lambda}(\cdot)$ can be soft-thresholding function or $\text{ReLU}(x - \lambda)$ to ensure sparsity of $\mathbf{S} = \sigma_{\lambda}(\attn \inp \coeff) \in \mathbb{R}^{N \times C_o}$, 
$\coeff \in \mathbb{R}^{C_i\times M}$ and $\atoms \in \mathbb{R}^{M \times C_o}$ is the feature dictionary atoms, $M$ is the number of atoms.

With this formulation, we obtain the feature dictionary atoms $\atoms$ (we use the term ``atoms" for simplification in this paper), which are the building blocks for feature representation $\out$. 
We illustrate our formulation in Figure~\ref{fig:method_ft} (a).
This formulation provides the following properties:
\begin{itemize}
    \item The sparse coefficients $\mathbf{S}=\sigma_{\lambda}(\attn \inp \coeff)$ ensure that the feature representation $\out$ is combined by only a few $\atoms$. We present an extreme example to illustrate how atoms enhance interpretability: $\out_i = \sum_{j=1}^M \mathbf{S}_{ij} \atoms_j$ can be simplified as $\out_i = \mathbf{S}_{ij^*} \atoms_{j^*}$ if all elements of $\mathbf{S}_{i}$ is 0 except the element at index $j^*$, and only $\atoms_{j^*}$ contributes to the output feature $\out_i$.
    \item The sparse coefficients $\mathbf{S}$ are data-dependent, while the feature dictionary atoms $\atoms$ are data-independent. While different inputs generate different coefficients, the output features are always constructed using the same feature dictionary atoms.
    \item We can adjust the size of the feature dictionary by varying $M$ to handle tasks with different difficulties. Large $M$ leads to an overcomplete dictionary while small $M$ produces an undercomplete dictionary.
\end{itemize}


 

\vspace{-5pt}
\paragraph{Fine-tuning via sparse dictionary learning.}
The above formulation frames feature representation through the lens of sparse coding. In the context of large model fine-tuning, we freeze the pre-trained parameters to make feature representations $\out$ of the pre-trained model fixed. Instead, we introduce a sparse coding perspective to encode the adapted feature representation $\Delta \out$, which is learned from downstream tasks.
Specifically, we have,
\begin{equation}
    \out + \Delta \out = \out + \sigma_{\lambda}(\attn \inp \coeff) \atoms.
\end{equation}
With this formulation, the pre-trained weights remain unchanged, preserving the capabilities of the pre-trained model. 
The adapted feature $\Delta \out$ is represented as a linear combination of feature dictionary atoms $\atoms$, using sparse coefficients $\mathbf{S} = \sigma_{\lambda}(\attn \inp \coeff)$. 
This formulation offers a framework for understanding how $\atoms$ contributes to the adapted features in the fine-tuned model.
We illustrate our method in Figure~\ref{fig:method_ft} (a).

\subsection{Analysis of Dictionary Atoms}
\label{sec:alyz_atoms}

\begin{figure}[t]
  \centering
 \begin{subfigure}[b]{0.47\textwidth}
     \centering
     \includegraphics[trim={250pt 90pt 260pt 130pt}, clip, width=\textwidth]{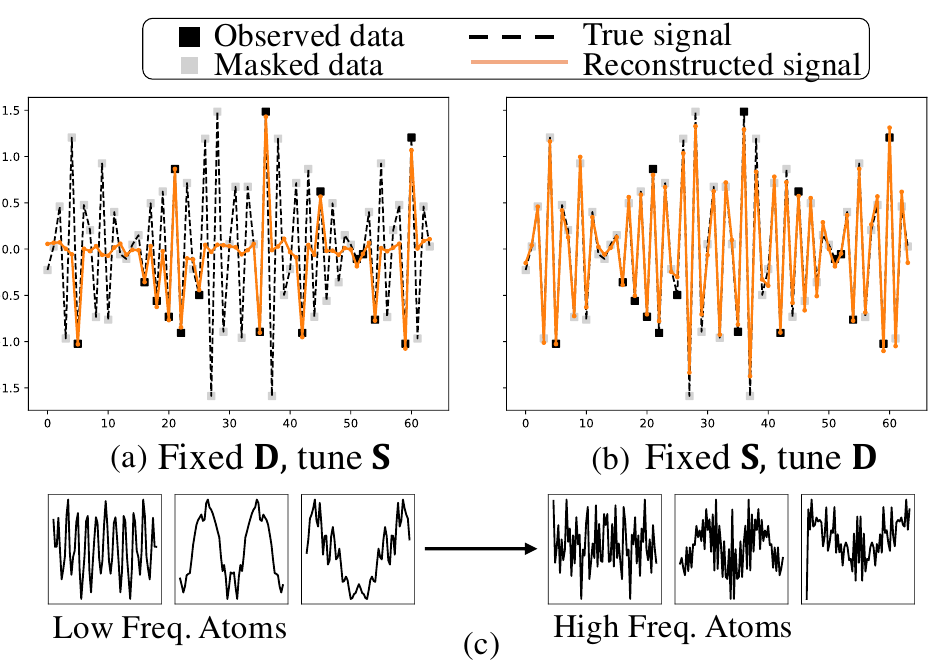}
 \end{subfigure}
 
  \caption{(a) Adjusting only the coefficients fails to adapt the model to the downstream task, whereas (b) modifying only the atoms enables a perfect fit to the new task composed of high-frequency signals. (c) Fine-tuning the atoms effectively transitions the model from low-frequency to high-frequency representations. They encode the representation of the downstream task.}
  \label{fig:toy2}
  \vspace{-6pt}
\end{figure}

In this section, we demonstrate distinct properties of atoms $\atoms$ and sparse coefficients $\mathbf{S}$.

\vspace{-5pt}
\paragraph{Features atoms are key to fine-tuning.}
We analyze the distinct behaviors of coefficients and atoms, and discover that atoms play a crucial role in adapting to new tasks.
To explore this within the context of an attention block, we perform a toy experiment where masked one-dimensional signals are reconstructed using an attention layer.
The signals, with a length of 64, are synthesized by selecting 5 random Fourier bases, with a low frequency ranging from $\frac{0}{64}$ to $\frac{24}{64}$. 
The objective is to reconstruct the signal after masking out $75\%$ of its values with only 16 randomly sampled observations. 
After we have a pre-trained model on the low-frequency signals,
we fine-tune the model to a new task where signals are constructed with high-frequency Fourier bases ranging from $\frac{25}{64}$ to $\frac{32}{64}$. 
In this experiment, 
we test two different cases:
(1) tuning only coefficients on the high-frequency data with atoms fixed, and (2) tuning only atoms on the high-frequency data with coefficients fixed.
As illustrated in Figure~\ref{fig:toy2}, when adapting to high-frequency signals, modifying only the atoms while keeping the coefficients fixed allows the model to effectively generalize to the new task. 
In contrast, updating only the coefficients is insufficient for proper adaptation.
Building on this observation, we hypothesize that adjusting atoms is more suitable than adjusting coefficients for adapting to a new task.
We observe from Figure~\ref{fig:toy2} (c) that atoms are capable of encoding the representation of the downstream task. The experiment details are in Appendix~\ref{appx:exp}.


 

\vspace{-5pt}
\paragraph{Identifying the importance of feature atoms.}
We analyze the impact of atoms on the newly learned concept by visualizing both the influence of individual atoms and their combined effect.
In this experiment, we adopt a transformer-based variational autoencoder (VAE)~\cite{kingma2019introduction} with a two-layer decoder. The experiment process is illustrated in Figure~\ref{fig:method_ft} (b).
We first pre-train the model on an image-denoising task using a subset of the MNIST dataset containing digits 5 to 9. We then fine-tune the model on a downstream denoising task with only digit 3 by learning a dictionary for each layer with $M=100$.
In this scenario, the digit ``3" is the newly learned concept.
Given the same noisy latent, the decoder outputs an ``8" without incorporating fine-tuned feature atoms, whereas it generates a ``3" when the fine-tuned feature atoms are applied.
The model requires only 12 atoms to shift the generated digits from 8 to 3. The influence of each atom is visualized in Figure~\ref{fig:method_ft} (c), with the overall impact represented as $\Delta \out$, where the red area indicates an increase in pixel value at a specific location and the blue area represents a decrease.
It is straightforward to see that these 12 feature atoms enhance the regions corresponding to ``3" while suppressing certain portions of ``8". 
Additionally, Figure~\ref{fig:method_ft} (c) illustrates the individual behavior of each atom, showing how different atoms affect distinct parts of the generated digit. 
The experiment details are in Appendix~\ref{appx:exp}.



\subsection{Design Variations}
In this section, we outline the various design choices implemented in our experiments.
Our method achieves sparse coefficient $\mathbf{S} = \sigma_{\lambda}(\attn \inp \coeff)$ by utilizing non-linear function $\sigma_{\lambda}(\cdot)$. In the literature on sparse coding~\cite{bengio2013representation}, there are three different methods: (i) soft-thresholding function, $\sigma_{\lambda}(\mathbf{x})=\text{prox}_{\lambda \|\cdot\|_1} (\mathbf{x}) = \text{sign}(\mathbf{x}) \odot \max(|\mathbf{x}| - \lambda, 0)$, (ii) ReLU, $\sigma_{\lambda}(\mathbf{x})=\text{ReLU} (\mathbf{x} - \lambda) = \max(\mathbf{x} - \lambda, 0)$, and (iii) top-$k$ activation selection, which selects the most significant features of top $k$ contributing to the learned representation.
In practice, we observe that top-$k$ activation selection maintains a consistent level of sparsity across different inputs. In contrast, methods (i) soft-thresholding function and (ii) ReLU exhibit sensitivity to input values, leading to variations in sparsity. However, methods (i) and (ii) offer control over the coefficient values, ensuring they stay above a specified threshold.

Furthermore, we find that when $\sigma_{\lambda}(\inp \coeff)$ is highly sparse, the resulting product $\attn \sigma_{\lambda}(\inp \coeff)$ also retains sparsity. In modern transformers, computation of $\attn$ and $\inp \Wv$ are performed separately. For simplicity in implementation, we adopt the formulation $\attn \sigma_{\lambda}(\inp \coeff)$.


\begin{figure*}[t]
  \centering
 \begin{subfigure}[b]{0.84\textwidth}
     \centering
     \includegraphics[trim={0pt 90pt 0pt 0pt}, clip, width=\textwidth]{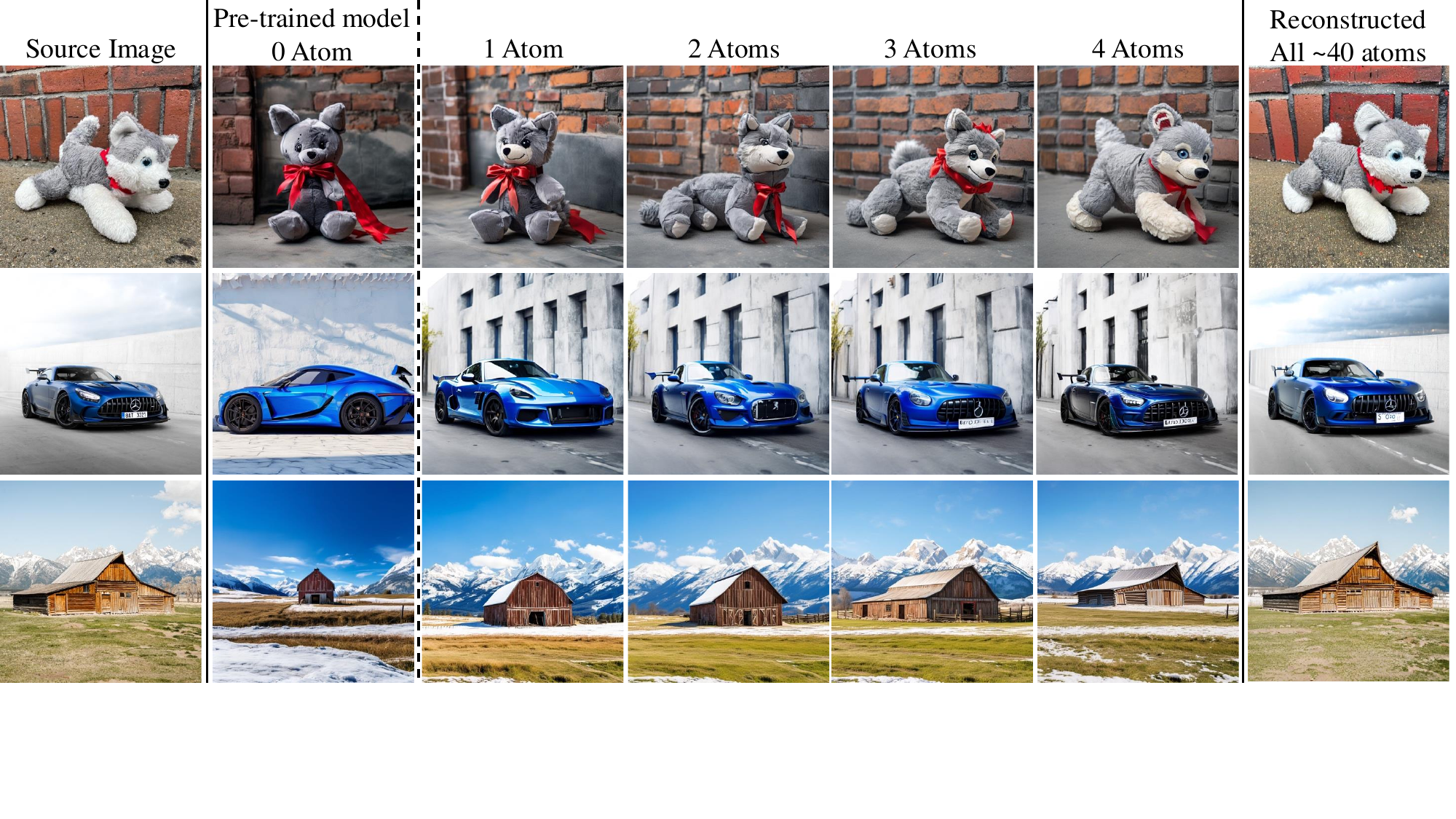}

 \end{subfigure}
 
  \caption{
  The pre-trained model can generate the target concept when integrated with the learned dictionary, even with about 40 atoms.
  With just four atoms, the core structure of the concept can still be generated. 
  Adjusting the number of selected atoms influences the generated results, offering a controllable approach to image editing.
  Additional discussions of these figures are in Section~\ref{exp:dit_atoms}.}
  \label{fig:exp_dit_atoms}
  \vspace{-8pt}
\end{figure*}

\subsection{Analysis of Our Design}
\label{sec:anlyz_lora}

\paragraph{Comparison with LoRA.}
When representing attention mechanism (\ref{eq:attn_map}-\ref{eq:multi_head_attn}) as a linear combination of dictionary atoms,
updating $\Wq, \Wk$ is equivalent to updating the coefficients while updating $\Wv, \Wo$ is equivalent to updating the atoms. 
LoRA applies adaptation on $\Wq$, $\Wk$, $\Wv$, $\Wo$ with a low-rank matrices $\mathbf{W}_A \in \mathbb{R}^{C_i \times r}, \mathbf{W}_B \in \mathbb{R}^{r \times C_o}$, where $r$ represents the rank. LoRA adapts both coefficients and atoms in a low-rank way.
In comparison, our method adapts the coefficients with $\coeff$ and obtains sparse coefficients $\sigma_{\lambda}(\attn \inp \coeff)$, while representing atoms with $\atoms$. 

\vspace{-7pt}
\paragraph{Parameters and computational cost.}
To adapt the representation at each layer, we need additional parameters for $\coeff$ and $\atoms$, which are $(C_i+C_o)M$. After the fine-tuning, considering the density of the coefficients $\rho$, i.e., $1-\rho$ is sparsity, the storage required for the parameters is $C_i M+\rho C_o M$. In comparison, LoRA requires $4(C_i+C_o)r$ additional parameters.
The FLOPs required by our method are $2NMC_i+2\rho NMC_o$. LoRA requires $8 C_i C_o r$ FLOPs.
For example, if $C_i=C_o=1024$, $M=256$, $r=16$, $N=1024$, $\rho=0.01$, our method requires $264$k parameters for storage and $0.54$GFLOPs for compututation, while LoRA requires $131$k parameters and $0.13$GFLops. Our method involves slightly higher computational costs than LoRA, as it accounts for feature interactions, whereas LoRA only updates the weights. Next, we demonstrate that by leveraging sparse feature representations, our approach achieves greater robustness compared to LoRA.

\begin{figure}[t]
  \centering
 \begin{subfigure}[b]{0.44\textwidth}
     \centering
     \includegraphics[trim={300pt 210pt 260pt 20pt}, clip, width=\textwidth]{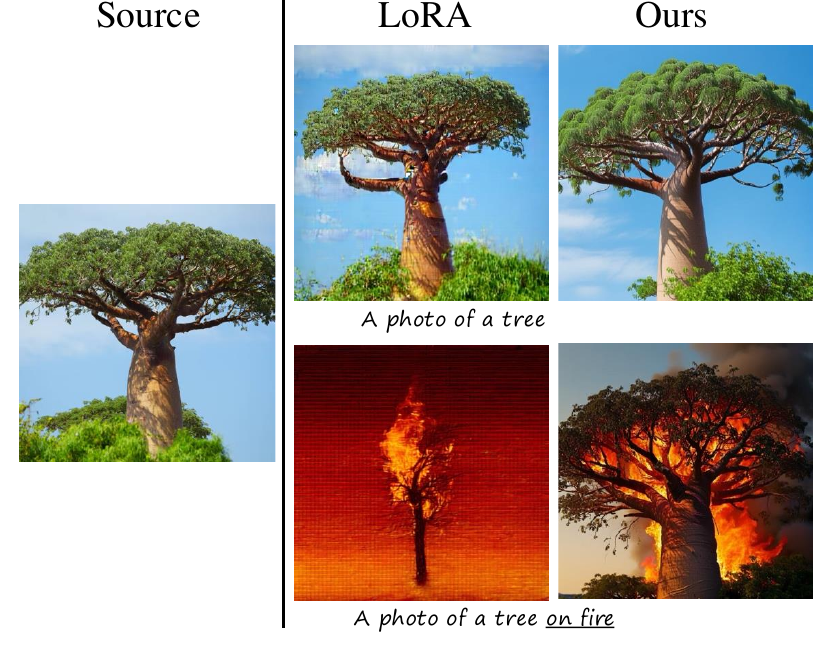}

 \end{subfigure}
 
  \caption{
  The role of sparsity: 
  Our method utilizes the same number of atoms as LoRA, yet LoRA struggles to adapt to the target prompt effectively. Sparse coding ensures that atoms are either activated or inactive, providing a structured and interpretable adaptation process. In contrast, LoRA produces noisy coefficients, leading to completely disrupted outputs.
  }
  \label{fig:exp_abla_lora}
  \vspace{-10pt}
\end{figure}

\vspace{-7pt}
\paragraph{Our method provides more stable adaptation.}
We compare an extreme case between LoRA and our method,
by considering the attention in a dictionary form $\Delta \out = \attn \mathbf{V}$ or $\Delta \out=\mathbf{S} \atoms$. 
In this experiment, we examine the difference between updating coefficients using a low-rank approach versus a sparse approach.
Both LoRA and our method produce the updated feature $\Delta \out$ with rank 1, where LoRA achieves this by setting $r=1$ and our method enforces only one active atom in $\atoms$. 
The key difference lies in how the coefficients are updated—LoRA adapts them in a low-rank manner, whereas our method updates them sparsely.
As shown in Figure~\ref{fig:exp_abla_lora}, by fine-tuning the model on a single concept, both LoRA and our method successfully reconstruct the concept using the same prompt. 
LoRA fails to produce meaningful images with a slightly modified prompt, while our method remains robust, consistently generating a concept that aligns with the updated text prompt.
We attribute the effectiveness of our approach to the sparse coefficients produced by $\sigma_{\lambda}(\attn \inp \coeff)$. With a different text prompt, the values in $\mathbf{S}$ either remain zero or take on a significant value when receiving the proper input, ensuring that the generated result remains coherent. In contrast, LoRA produces noisy coefficients, leading to completely disrupted outputs.

\section{Experiments}

In this section, we evaluate the effectiveness of our method in image generation tasks.
We demonstrate that feature dictionary atoms effectively represent the target concept after fine-tuning the pre-trained model. Compared to baselines, our approach enables selective atom adjustment, improving image editing by refining results and enhancing text-to-image concept customization while preserving pre-trained model capabilities.

\subsection{Representing Concepts Using Atoms}
\label{exp:dit_atoms}

\paragraph{Experimental settings.}
In this experiment, we utilize PixArt–$\Sigma$~\cite{chen2024pixart}, a Diffusion Transformer (DiT)~\cite{peebles2023scalable} model consisting of 28 transformer blocks. We fine-tune the model on a single image for 60 epochs using our method and reconstruct the corresponding concept using the same prompt used during fine-tuning. The selected images are drawn from DreamBooth~\cite{ruiz2023dreambooth}, Custom Diffusion~\cite{kumari2023multi}.
We control the number of active atoms and visualize the variation of the generated results with different atoms.

\vspace{-5pt}
\paragraph{The concept is represented by a set of atoms.}
As shown in Figure~\ref{fig:exp_dit_atoms},
after fine-tuning our method on the target concept, its representation can be effectively captured using a small set of atoms, such as 40 atoms. Among these, just 4 atoms are sufficient to construct the core structure of the concept, including key attributes like pose and color. 
We provide details descriptions of these images in Appendix~\ref{appx:exp_results}.

\subsection{Image Editing}

\begin{figure}[t]
  \centering
 \begin{subfigure}[b]{0.44\textwidth}
     \centering
     \includegraphics[trim={300pt 70pt 260pt 20pt}, clip, width=\textwidth]{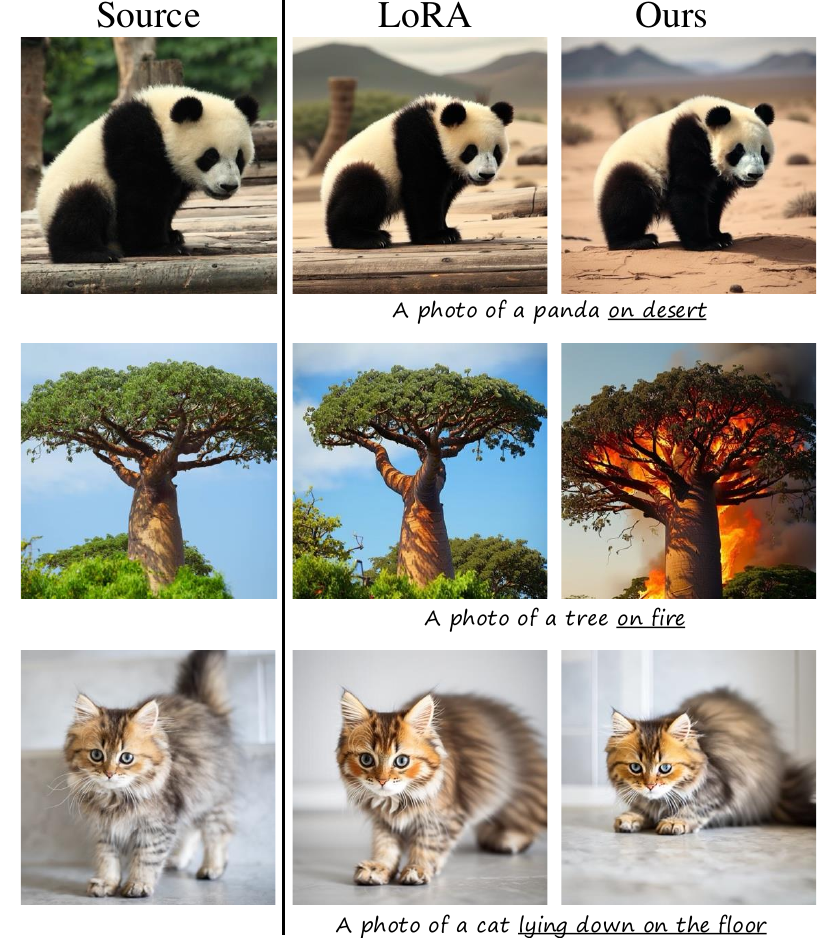}

 \end{subfigure}
 
  \caption{Results on image editing. Our method shows improved results by better following editing prompts while preserving the key components of the source images.}
  \label{fig:exp_editing}
\vspace{-5pt}
\end{figure}

Building on our previous observation that adjusting the number of atoms influences the generated results, we explore leveraging this property for image editing.
Image editing tasks aim to modify an image using a generative model while retaining as many details as possible from the original~\cite{zhang2023sine}. 
However, certain editing tasks pose challenges in maintaining most of the original details while achieving effective modifications. For example, altering the style can significantly change the overall appearance of the image. We argue that successful image editing should preserve the main components of the image while applying modifications selectively. Since our method represents the core components of a concept using only a few atoms, it is well-suited for image editing tasks, ensuring that key structural elements remain intact while allowing for controlled transformations.

\paragraph{Experimental settings.}
In our experiment, we select data from previous work~\cite{zhang2023sine} and evaluate different editing tasks, including background modification, scene alteration, and pose adaptation. 
Previous works~\cite{han2023svdiff, zhang2023sine} fine-tune models to learn a concept and modify its appearance by adjusting the prompt. However, these approaches have not yet been applied to DiT. Following their setup, we compare our method with fine-tuning approaches, such as LoRA~\cite{hu2021lora}, using them as baselines for evaluation.
We adopt the DiT architecture and fine-tune both LoRA and our method on a single image for 60 epochs to achieve full reconstruction. After fine-tuning, we modify the image using a different prompt for editing.

\begin{figure}[t]
  \centering
 \begin{subfigure}[b]{0.44\textwidth}
     \centering
     \includegraphics[trim={330pt 210pt 180pt 150pt}, clip, width=\textwidth]{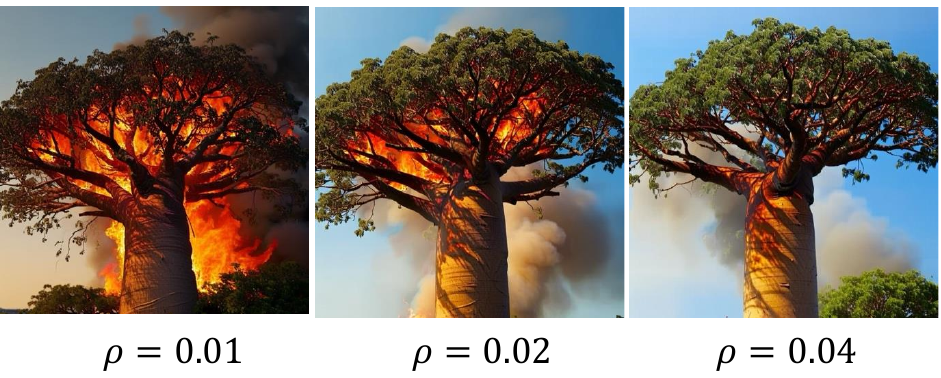}

 \end{subfigure}
 
  \caption{Adjusting the density, \textit{i.e.}, the number of activated atoms, can change the influence on the edited image.}
  \label{fig:exp_adjust}
  \vspace{-5pt}
\end{figure}

\paragraph{Image editing results.}
The results are shown in Figure~\ref{fig:exp_editing}. 
Our method applies edits based on the test prompt while maintaining the core structure of the source images.
In addtion, we control the number of active atoms by adjusting the coefficient density $\rho$, where $1-\rho$ represents the sparsity level.
In Figure~\ref{fig:exp_adjust}, as sparsity increases, only the most essential atoms remain, preserving the core concept while allowing the pre-trained model greater flexibility to integrate editing effects based on the modified text prompt.

\subsection{Concept Customization}
In the previous section, we demonstrated that our method can efficiently represent the target concept using only a small number of atoms. In this section, we further show that even with just a few atoms, our method achieves good performance in concept customization.

\paragraph{Experimental settings.}
We evaluated our approach on the DreamBooth~\cite{ruiz2023dreambooth} dataset, which consists of 30 subjects across 15 distinct classes. Among these, 9 are live subjects, while the remaining 21 are objects. Each subject has between 4 to 6 images, captured under varying conditions, in diverse environments, and from multiple angles. All 30 subjects were trained for 15 epochs and validated on 25 different prompts.
For our experiments, we adopted the Pixart-$\Sigma$, a Diffusion Transformer (DiT)~\cite{chen2024pixart}, as the base architecture and fine-tuned it using our proposed method. To verify the effectiveness of our approach, we compared its performance against several PEFT baselines for personalized text-to-image generation, including LoRA~\cite{hu2021lora}, DoRA~\cite{liu2024dora}, and OFT~\cite{oft}. 
We use the AdamW optimizer with a learning rate of $1 \times 10^{-4}$ to fine-tune the Pixart-$\Sigma$~\cite{chen2024pixart} for 15 steps. For the baseline methods, LoRA and DoRA are assigned a rank of $r = 16$, while OFT is set to $r = 4$. We generated five images for each prompt at a resolution of $1024 \times 1024$.

\paragraph{Evaluation metrics.}
We measure concept alignment by computing the mean cosine similarity between the generated image’s CLIP embeddings~\cite{radford2021learning} and those of the original concept images. We quantify image diversity using the Vendi score~\cite{friedman2022vendi} with DINOv2 embeddings~\cite{oquab2023dinov2}. For text alignment, we calculate the average cosine similarity in the CLIP feature space~\cite{radford2021learning}. ImageReward~\cite{xu2023imagereward} is a general-purpose text-to-image human preference reward model that evaluates and ranks AI-generated images by scoring them based on human preferences. 

\paragraph{Customization results.}
As shown in Table~\ref{exp:concept}, compared with baseline methods, our approach not only represents the concept with a minimal set of learned atoms but also generalizes robustly to diverse scenarios. In particular, our method achieves higher text-to-image alignment and diversity scores than most other PEFT baselines, while still preserving a relatively high fidelity score. This indicates that our approach effectively captures essential features of each concept without sacrificing the quality of customization text-to-image generation.

\paragraph{Number of parameters in different methods.}

We compare the trainable parameters of our method and the baselines. The pre-trained PixArt model has around 616M parameters in total. Among the fine-tuning methods, both LoRA and DoRA have around 4M trainable parameters. OFT uses around 37M trainable parameters, and our method trains around 17M parameters.

\begin{table}[]
\caption{Comparison with baseline methods on concept customization.}
\centering
\scalebox{0.8}{
\begin{tabular}{c|ccccc}
\toprule
                & Alignment & Diversity & Fidelity & ImageReward \\ \midrule
LoRA~\cite{hu2021lora} & 0.218     & 4.707     & 0.711    & \underline{0.220} \\
DoRA~\cite{liu2024dora} & 0.213    & 4.540    & \underline{0.712}    & 0.142 \\
OFT~\cite{oft}         & 0.190     & 3.926     & 0.663    & 0.137  \\ \midrule
Ours ($\rho=0.02$)   & \textbf{0.235}     & \textbf{5.205}     & 0.702    & 0.182 \\
Ours ($\rho=0.04$)   & \underline{0.231}     & \underline{5.172}     & \textbf{0.713}    & \textbf{0.431}  \\
\bottomrule
\end{tabular}
}
\label{exp:concept}
\end{table}


\subsection{Abalation Study}
\begin{figure}[t]
  \centering
 \begin{subfigure}[b]{0.47\textwidth}
     \centering
     \includegraphics[trim={10pt 10pt 10pt 10pt}, clip, width=\textwidth]{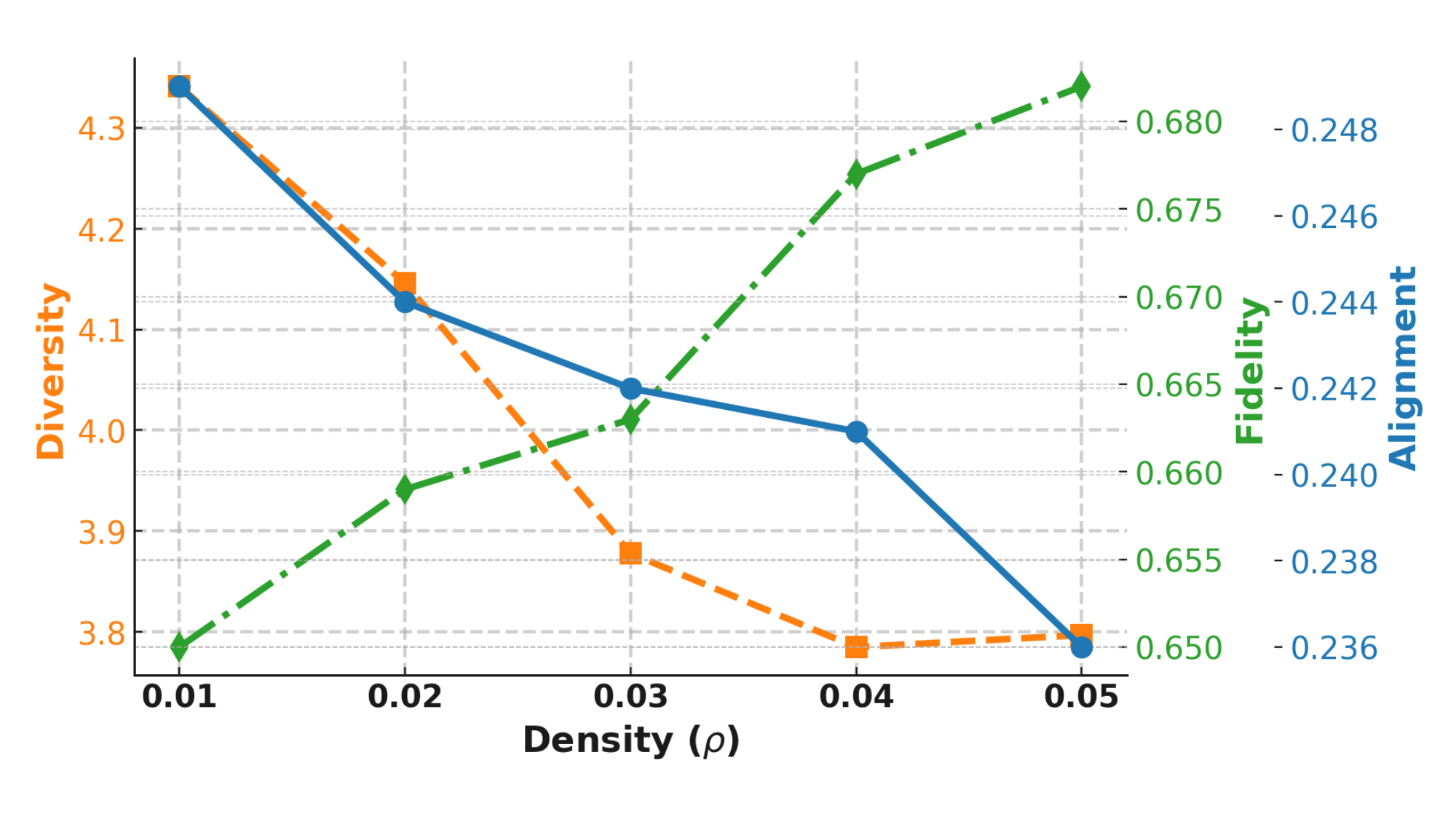}

 \end{subfigure}
 
  \caption{The influence of different density $\rho$ on the image diversity, fidelity and text-to-image alignment.}
  \label{fig:rho_influence}
\end{figure}

\paragraph{The impact of sparsity in the coefficients.}
We analyze the impact of varying sparsity levels in the coefficients through a concept customization experiment, evaluating text alignment, image diversity, and image fidelity. As shown in Figure~\ref{fig:rho_influence}, higher sparsity (low density) enhances text alignment and image diversity but results in slightly lower image fidelity. In contrast, lower sparsity (high density) improves fidelity, indicating that using more atoms allows for a more precise representation of the concept from the source image. In our experiment, we adopt both $\rho=0.02$ and $\rho=0.04$ for either better diversity or better fidelity.


\vspace{-5pt}
\paragraph{The impact of member numbers $M$.} 
We also examine the impact of the number of members, \textit{i.e.}, the dictionary size, on the generated results while maintaining the same level of sparsity. As shown in Table~\ref{tab:abla_member}, increasing the number of members improves image fidelity, whereas a smaller dictionary size reduces fidelity but enhances text alignment and image diversity. In our experiment, we choose $M=256$ to provide balanced fidelity and diversity.

\begin{table}[]
\centering
\caption{The influence of different member numbers $M$. As the number of members increases, alignment and diversity scores decrease, while the fidelity score improves.}
\scalebox{0.8}{
\begin{tabular}{c|ccc}
\toprule
$\rho=0.01$        & Alignment & Diversity & Fidelity \\ \midrule
M = 16    & \underline{0.248}     & \textbf{5.867}     & 0.547    \\
M = 256   & \textbf{0.249}     & \underline{4.341}     & 0.650    \\
M = 1024  & 0.241     & 3.733     & \underline{0.672}    \\
M = 2048  & 0.238     & 3.525     & \textbf{0.692}    \\ \bottomrule
\end{tabular}
}
\label{tab:abla_member}
\vspace{-5pt}
\end{table}


\vspace{-5pt}
\paragraph{Sparsity coefficients provide more stable adaptation.}
In this experiment, we compare our method with LoRA on the image editing task. As we discussed in Section~\ref{sec:anlyz_lora}, LoRA adapts the coefficients in a low-rank manner, whereas our method updates them sparsely. Both LoRA and our method produce the updated feature $\Delta \out$ with rank 1, where LoRA achieves this by setting $r=1$ and our method enforces only one active atom in $\atoms$. 
As shown in Figure~\ref{fig:exp_abla_lora}, by fine-tuning the model on a single concept, both LoRA and our method successfully reconstruct the concept using the same prompt. However, when applying the fine-tuned weights to generate images with a slightly modified prompt, LoRA becomes unstable, failing to produce meaningful images. In contrast, our method remains robust, consistently generating a concept that aligns with the updated text prompt.


\section{Conclusion}
In this paper, we introduce to frame model fine-tuning as a sparse coding problem, where the adapted feature representation is constructed using a dictionary of feature atoms. We propose a fine-tuning framework that leverages a combination of feature dictionary atoms, demonstrating that tuning them enhances control over the adaptation process.
We validate the effectiveness of our approach across various generative tasks. By representing concepts with only a few essential atoms, our method proves to be highly adaptable and can be effectively applied to image editing, style mixing, and concept personalization.
Our study provides a new perspective on the mechanics of fine-tuning, paving the way for more controllable adaptation strategies in large-scale pre-trained models.

{
    \small
    \bibliographystyle{ieeenat_fullname}
    \bibliography{main}
}

\clearpage
\setcounter{page}{1}
\maketitlesupplementary
\renewcommand{\thesection}{\Alph{section}}
\setcounter{section}{0}

\section{Analysis}
\label{appx:alys}

\subsection{Analysis of Adapted Feature Representation}
With our formulation, the adapted feature representation $\Delta \out$ at each layer can be expressed as a linear combination of learned atoms $\atoms$. However, deep neural networks represent features non-linearly. Although $\Delta \out$ can be expressed linearly in terms of $\atoms$, its influence becomes non-linear as it interacts with activation functions and transformations in subsequent layers. 
In this section, we introduce an approximation method to address this non-linearity, enabling us to trace and quantify the contribution of each atom to the final outputs throughout the entire model.

\paragraph{Formulation of the pre-trained model.}
We represent each layer of the pre-trained model as $f^{(l)}(\mathbf{x})$. For an $L$-layer model, it writes as $f^{(L)} \circ \cdots \circ f^{(1)}(\mathbf{x})$, where $l=1, \cdots, L$ is the index of the layer. 
Based on the functional approximation theory, we can approximate the function in an appropriate basis. Given a basis $\{g_k\}_{k=1}^{K}$, the coefficients of basis expansion are $\mu_k=\langle f,g_k\rangle$, such that $f(\mathbf{x}) = \sum_{k=1}^{K} \mu_k g_k(\mathbf{x})$, where $\langle\cdot, \cdot \rangle$ is inner product. For simplified analysis, we utilize the polynomial basis,
and represent $f(\mathbf{x})$ as:
\begin{equation}
    f(\mathbf{x}) = \sum_{k=1}^{K} c_k \mathbf{x}^{k}.
    \label{eq:poly}
\end{equation}

\paragraph{The influence of atoms on one layer.}
Fine-tuning introduces adapted feature representation at layer $l-1$, \textit{i.e.}, $\Delta \out^{(l-1)}=\mathbf{S} \atoms$, which become the perturbation of the input $\inp^{(l)}$ at layer $l$. 
For each input, we have $\mathbf{x} + \sum_{m=1}^M s_m \mathbf{d}_m$, where $\mathbf{d}_m \in \mathbb{R}^{C_o}$ is the single atom in $\atoms$ and $s_m$ is the corresponding sparse coefficient. With the perturbation, we have
\begin{equation}
    f(\mathbf{x} + \sum_{m=1}^M s_m \mathbf{d}_m) 
    = f(\mathbf{x}) + \sum_{m=1}^M \mathbf{B}_{K}(\mathbf{d}_m),
    \label{eq:2layer}
\end{equation}
where $\mathbf{B}_{K}(\mathbf{d}_m)= \sum_{k=1}^{K} \beta_{k, m} \mathbf{d}_m^k$ and $\beta_{k, m}$ depends on $\mathbf{x}$, $s_m$, and $c_k$. We assume $\langle \mathbf{d}_i, \mathbf{d}_j \rangle=0, \forall i \neq j$ to avoid correlated influence from different atoms. It can be achieved with a simple regularization term. 
With the influence of linear perturbations based on atoms at layer $l-1$, the different in the output at layer $l$ is determined by the linear combination of higher-order terms of these atoms $\mathbf{d}_m^k$.

\paragraph{The accumulated influence of atoms on multiple layers.}
Applying the same formulation to layer $l+1$ results in a similar expression as above, but incorporates the combined influence of atoms from both layer $l-1$ and layer $l$. Specifically, the input of layer $l+1$ is written as,
\begin{equation}
    \mathbf{x}^{(l+1)} + \sum_{m=1}^M s_m^{(l)} \mathbf{d}_m^{(l)} + \sum_{m=1}^M \mathbf{B}_{K^{(l-1)}}(\mathbf{d}_m^{(l-1)}),
\end{equation}
where $\mathbf{x}^{(l+1)}$ is the original input at layer $l+1$, $\sum_{m=1}^M s_m^{(l)} \mathbf{d}_m^{(l)}$ is the linear perturbations based on the atoms from layer $l$, $\sum_{m=1}^M \mathbf{B}_{K^{(l-1)}}(\mathbf{d}_m^{(l-1)})$ is the influence of the atoms from layer $l-1$, which is a combination of higher-order terms.
The output of layer $l+1$ is,
\begin{equation}
\begin{split}
    & f^{(l+1)}(\mathbf{x}^{(l+1)} + \sum_{m=1}^M s_m^{(l)} \mathbf{d}_m^{(l)} + \sum_{m=1}^M \mathbf{B}_{K^{(l-1)}}(\mathbf{d}_m^{(l-1)})) \\ 
    = & f(\mathbf{x}) + \sum_{m=1}^{M} \mathbf{B}_{K^{(l)}}(\mathbf{d}_m^{(l)}) + \sum_{m=1}^{M} \mathbf{B}_{K^{(l-1)}+K^{(l)}}(\mathbf{d}_m^{(l-1)}).
    \label{eq:3layer}
\end{split}
\end{equation}
This formulation can be naturally extended to the whole model. 
We provide detailed analysis in Appendix~\ref{appx:alys}.

After incorporating atoms into each layer to linearly capture the adapted feature representations, the model output can be interpreted as the original pre-trained output with perturbations introduced by the atoms at each layer. 
The atoms from the shallow layer (e.g., layer $1$), introduce more higher-order terms to the final output, i.e., $\mathbf{B}_{\sum_{l=1}^{L} K^{(l)}}(\mathbf{d}_m^{(1)})$. It contributes more to the overall structure of the output. The atoms from the deep layer ((e.g., layer $L$)), introduce fewer higher-order terms, or only linear terms to the final output. It contributes more to the details of the output. 

\subsection{Proof}

\paragraph{The proof of (\ref{eq:2layer}).}
To prove (\ref{eq:2layer}), we assume $\langle \mathbf{d}_i, \mathbf{d}_j \rangle=0, \forall i \neq j$ to avoid correlated influence from different atoms. It can be achieved with a simple regularization term. 
\begin{proof}
Inserting (\ref{eq:poly}) to the LHS of (\ref{eq:2layer}), we have
\begin{equation}
    f(\mathbf{x} + \sum_{m=1}^M s_m \mathbf{d}_m) = \sum_{k=1}^{K} c_k (\mathbf{x} + \sum_{m=1}^M s_m \mathbf{d}_m)^k.
\end{equation}
After expanding this equation, we have
\begin{equation}
\begin{split}
    & \sum_{k=1}^{K} c_k (\mathbf{x} + \sum_{m=1}^M s_m \mathbf{d}_m)^k \\
    = & \sum_{k=1}^{K} c_k \left[ \sum_{j=0}^k \begin{pmatrix} k \\ j\end{pmatrix}\mathbf{x}^{k-j} \left(\sum_{m=1}^M s_m \mathbf{d}_m \right)^j \right].
\end{split}
\end{equation}
Considering $\langle \mathbf{d}_i, \mathbf{d}_j \rangle=0, \forall i \neq j$, we have, 
\begin{equation}
    \left(\sum_{m=1}^M s_m \mathbf{d}_m \right)^j = \sum_{m=1}^M \left(s_m \mathbf{d}_m \right)^j.
\end{equation}
Thus,
\begin{equation}
\begin{split}
    & f(\mathbf{x} + \sum_{m=1}^M s_m \mathbf{d}_m) \\
    =& \sum_{k=1}^{K} c_k \left[ \mathbf{x}^{k} + \sum_{j=1}^k \begin{pmatrix} k \\ j\end{pmatrix}\mathbf{x}^{k-j} \sum_{m=1}^M \left(s_m \mathbf{d}_m \right)^j \right]\\
    =& f(\mathbf{x}) 
    + \sum_{m=1}^M \sum_{k=1}^{K} c_k \sum_{j=1}^{k} \begin{pmatrix} k \\ j\end{pmatrix}\mathbf{x}^{k-j} \left(s_m \mathbf{d}_m \right)^j. 
\end{split}
\end{equation}
We define 
\begin{equation}
    \mathbf{B}_{K}(\mathbf{d}_m) = \sum_{k=1}^{K} c_k \sum_{j=1}^{k} \begin{pmatrix} k \\ j\end{pmatrix}\mathbf{x}^{k-j} \left(s_m \mathbf{d}_m \right)^j,
\end{equation}
which represents the higher-order terms of $\mathbf{d}_m$ up to order $K$.
\end{proof}

\paragraph{The proof of (\ref{eq:3layer}).}
Here we only provide a sketch of proof of (\ref{eq:3layer}). Compared with (\ref{eq:2layer}), (\ref{eq:3layer}) contains an additional term $\sum_{m=1}^M \mathbf{B}_{K^{(l-1)}}(\mathbf{d}_m^{(l-1)})$. After applying the expansion (\ref{eq:poly}), we will have 
\begin{equation}
    \sum_{j=0}^k \begin{pmatrix} k \\ j\end{pmatrix}(\mathbf{B}_{K^{(l-1)}}(\mathbf{d}_m^{(l-1)}))^{k-j} \left(\sum_{m=1}^M s_m \mathbf{d}_m \right)^j,
\end{equation}
which produces the higher-order terms of $\mathbf{d}_m^{(l-1)}$ up to order $K^{(l-1)} + K^{(l)}$, thus we have $\mathbf{B}_{K^{(l-1)}+K^{(l)}}(\mathbf{d}_m^{(l-1)})$ in (\ref{eq:3layer}).





\section{Experimental Settings}
\label{appx:exp}


\paragraph{MNIST generation with VAE.}
In this experiment, the input of MNIST has the size of $28 \times 28$. The VAE consists with an encoder and decoder, each with two layers, a feature dimension of 128, and four attention heads. It first reshape the input with a patch size of 7. The latent space is mapped to a dimension of 32. We train the model using the Adam optimizer with a learning rate of 0.001 for 20 epochs. 

\paragraph{Generative tasks with DiT.}
In this experiment, we use the CAME~\cite{luo2023came} optimizer with a learning rate of $1 \times 10^{-4}$ to fine-tune the Pixart-$\Sigma$~\cite{chen2024pixart}. For the baseline methods, LoRA and DoRA are assigned a rank of $r = 16$, while OFT is set to $r = 4$. We generated five images for each prompt at a resolution of $1024 \times 1024$. 
We run the experiment on Nvidia A5000 with 24GB memory. For the image editing task, we train the model on 1 image for 60 epochs, which takes about 5min. For the concept customization task, we train the model on 4-6 images for 15 epochs, which takes about 5min.

\section{Additional Experimental Results}
\label{appx:exp_results}

\subsection{Toy Experiment}
\begin{figure}[t]
  \centering
 \begin{subfigure}[b]{0.47\textwidth}
     \centering
     \includegraphics[trim={0pt 0pt 0pt 0pt}, clip, width=\textwidth]{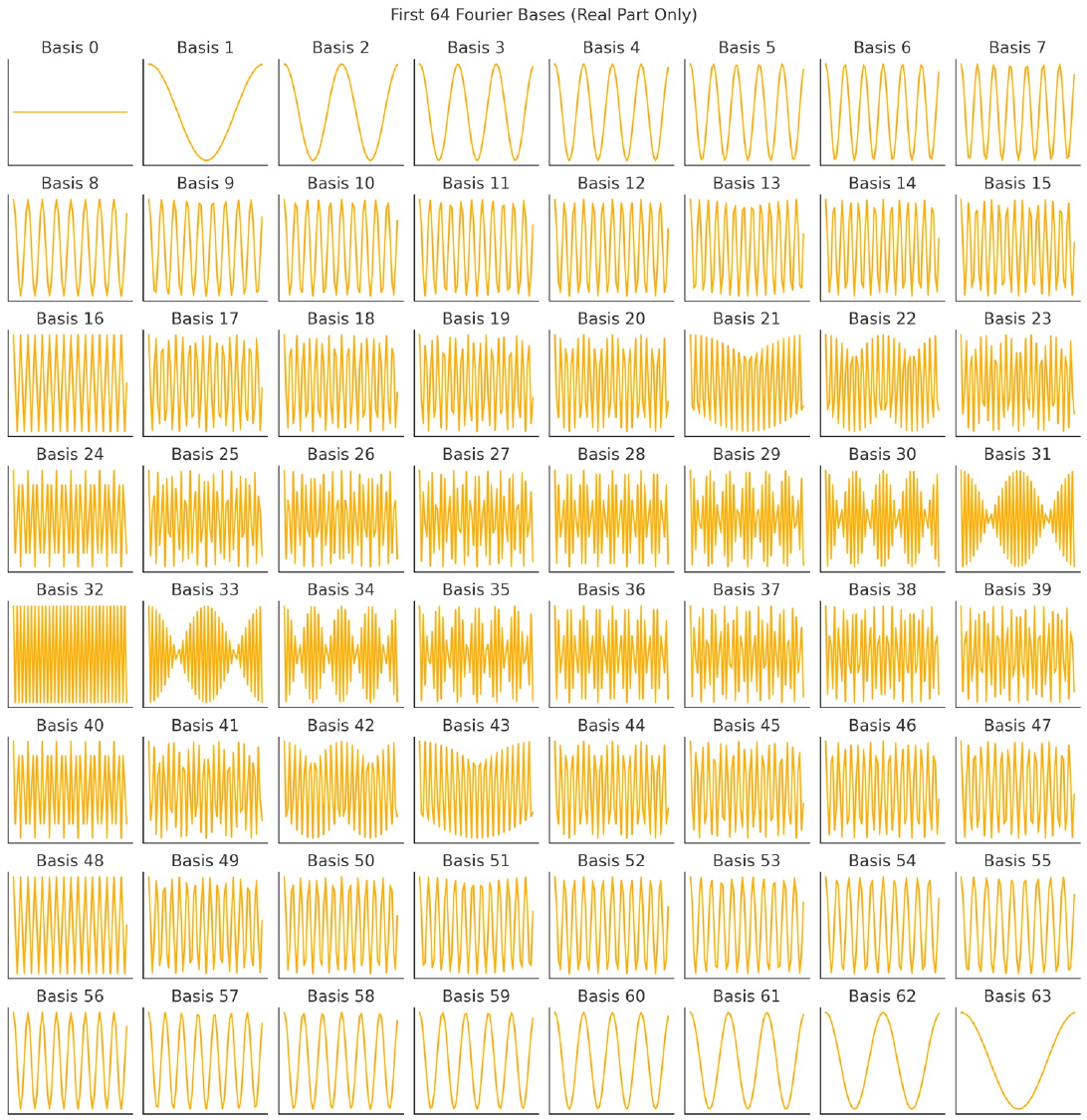}
 \end{subfigure}
 
  \caption{The illustration of Fourier basis.}
  \label{fig:fourier}
\end{figure}

\paragraph{Experimental setting.}
In this experiment, we only use the attention block to transform the signal, the feature dimension is 128, and the sequence length is 64. The synthetic signals are generated by randomly combining 5 Fourier bases, which is shown in Figure~\ref{fig:fourier}. To leverage the transformer, we first project the 1D signal into a 64D space using a randomly initialized projection matrix, which remains frozen during training. The output signal is then mapped from 64D back to 1D by summing across all 64 dimensions.

\paragraph{Sparse coefficients provide interpretability.}

\begin{figure}[t]
  \centering
 \begin{subfigure}[b]{0.47\textwidth}
     \centering
     \includegraphics[trim={290pt 50pt 280pt 40pt}, clip, width=\textwidth]{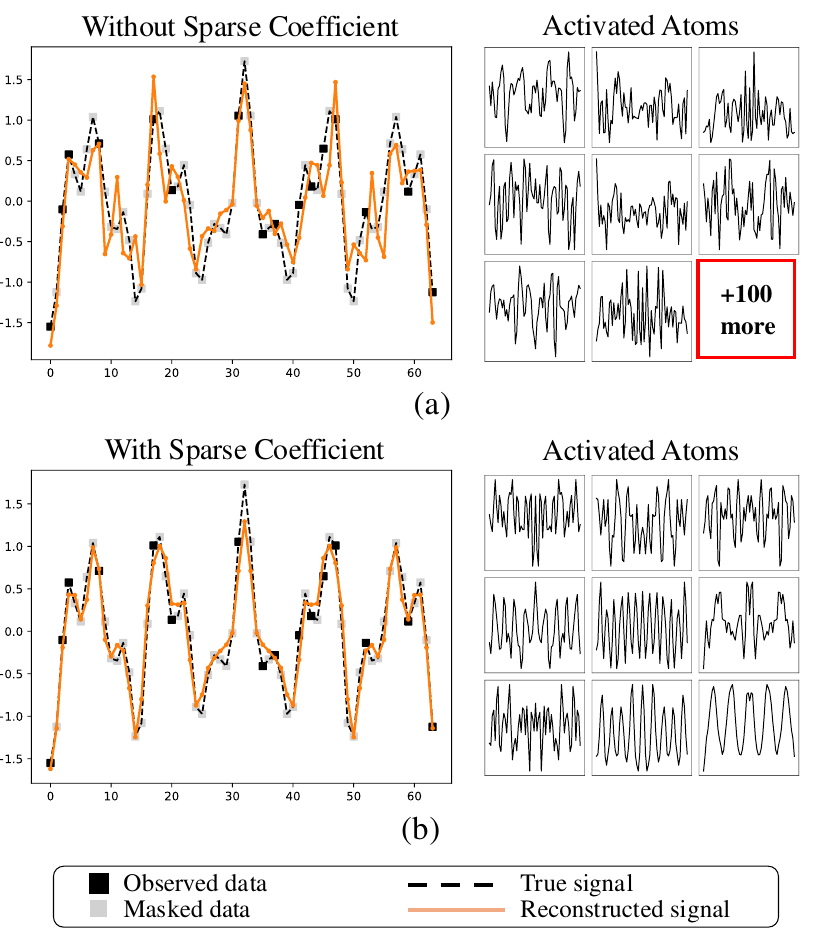}
 \end{subfigure}
 
  \caption{Compared to the performance of (a) standard attention $\attn \inp \Wv$, (b) the sparse combination of atoms $\sigma_{\lambda}(\attn \inp \coeff) \atoms$ provides more interpretability. The reconstructed signal is a linear combination of basic signals. Without sparse coefficients, constructing the signal requires 100 atoms. However, with sparse coefficients, the reconstructed signal is effectively represented using only 9 atoms, demonstrating a more compact and interpretable composition.}
  \label{fig:toy}
\end{figure}

The sparse coefficients $\mathbf{S}$ enhance interpretability by establishing a direct connection between a small subset of atoms and the output feature, \textit{i.e.}, $\out_i = \sum_{j=1}^M \mathbf{S}_{ij} \atoms_j$.
This behavior is also validated in the literature on sparse coding, compressed sensing, and dictionary learning~\cite{candes2006compressive}.
To explore this within the context of an attention block, we perform an experiment where masked one-dimensional signals are reconstructed using an attention mechanism.
The signals, with a length of 64, are synthesized by selecting 5 random Fourier bases, with a low frequency ranging from $\frac{0}{64}$ to $\frac{24}{64}$. 
The objective is to reconstruct the signal after masking out $75\%$ of its values with only 16 randomly sampled observations.  
As illustrated in Figure~\ref{fig:toy}, for both $\out = \attn \inp \Wv$ and $\out = \sigma_{\lambda}(\attn \inp \coeff) \atoms$, the attention block successfully reconstructs signals after jointly learning the coefficients and atoms. 
$\sigma_{\lambda}(\attn \inp \coeff) \atoms$ takes advantage of sparse coding, and requires only 9 atoms to fully reconstruct the signal. In comparison, $\attn \inp \Wv$ requires 100 atoms to construct the signal.
Interestingly, the atoms $\atoms$ in $\sigma_{\lambda}(\attn \inp \coeff) \atoms$ naturally resemble certain Fourier basis functions from the ones used to synthesize real data.

\paragraph{Description of reconstructed images.}
Figure~\ref{fig:exp_dit_atoms} also shows that the number of selected atoms impacts the generated results. 
For example, when learning the concept described in the top row, ``\textit{A grey $\langle V\rangle$ wolf plushie ...}", by adjusting the number of active atoms we can see that some atoms influence the texture of the fur, while others shape the posture of the object or determine the position of the bow tie. 
For the concept in the second row, ``\textit{A blue $\langle V\rangle$ sports car ...}", we observe that certain atoms influence the orientation of the sports car, while others affect the position of the rear wing and the shape of the grille. 
For the concept in the third row, ``\textit{A $\langle V\rangle$ wooden barn  ...}", we observe that certain atoms influence the number of lean-tos on the barn, while others affect the slope of the roof.

\subsection{Personalization Results Comparison}
In this section, we showcase the comparison of personalization results for various concepts selected from the DreamBooth~\cite{ruiz2023dreambooth} dataset. 

\begin{figure*}[t]
  \centering
 \begin{subfigure}[b]{1\textwidth}
     \centering
     \includegraphics[trim={0pt 10pt 20pt 20pt}, clip, width=\textwidth]{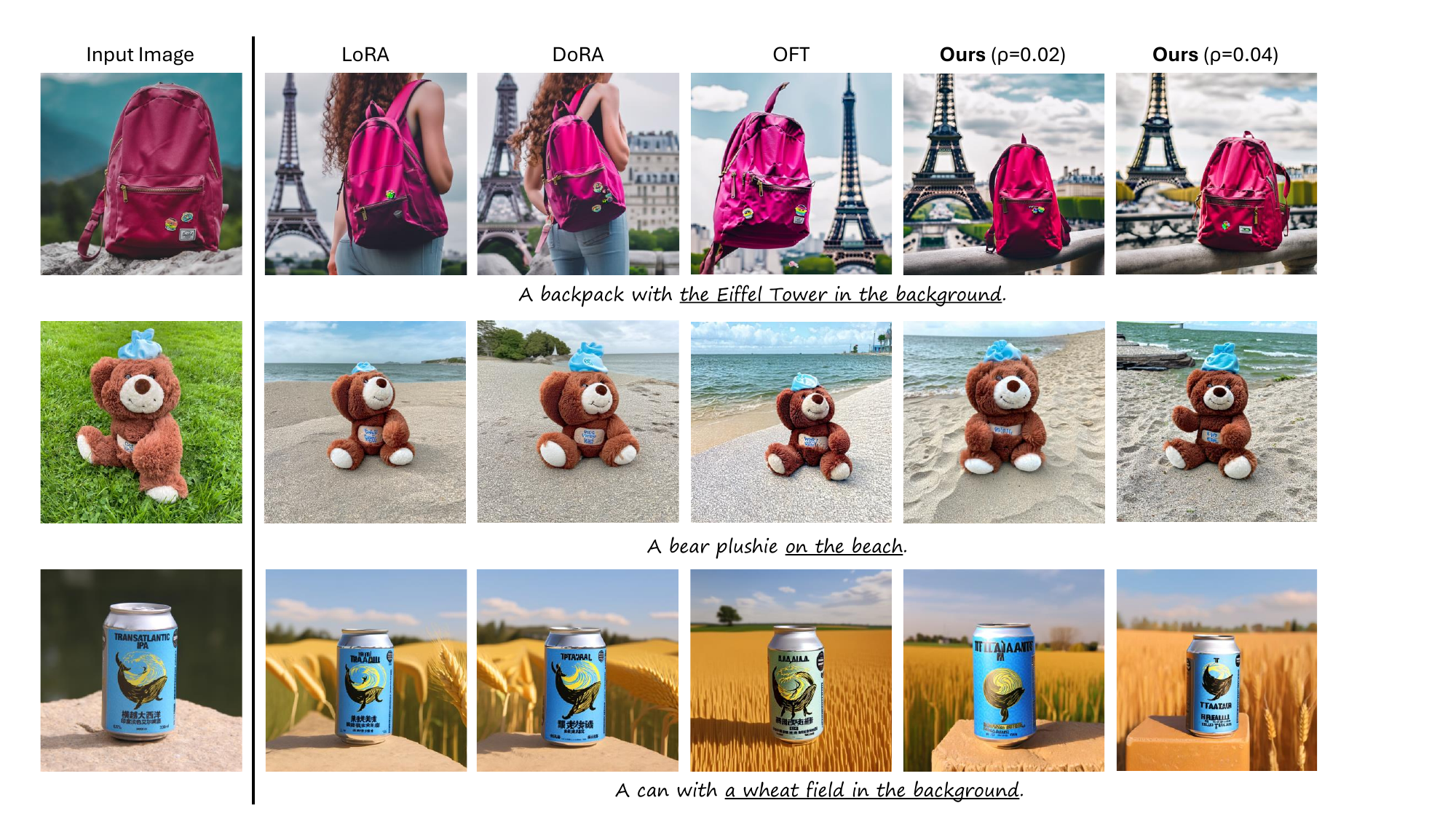}

 \end{subfigure}
 
  \caption{Personalization generated results comparison}
  \label{fig:personalization_1}
\end{figure*}

\begin{figure*}[t]
  \centering
 \begin{subfigure}[b]{1\textwidth}
     \centering
     \includegraphics[trim={0pt 10pt 20pt 20pt}, clip, width=\textwidth]{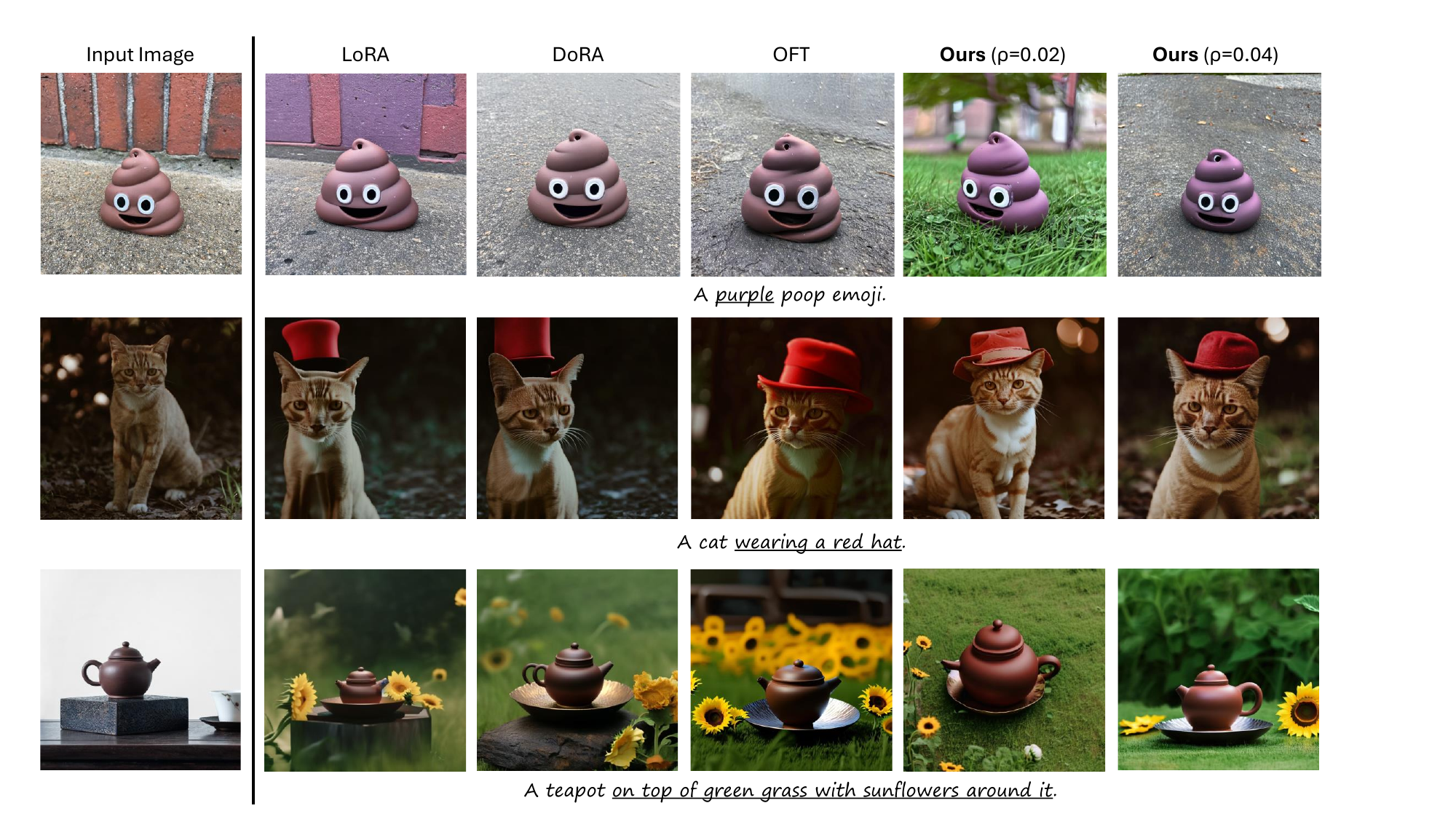}

 \end{subfigure}
 
  \caption{Personalization generated results comparison}
  \label{fig:personalization_2}
\end{figure*}

\begin{figure*}[t]
  \centering
 \begin{subfigure}[b]{1\textwidth}
     \centering
     \includegraphics[trim={0pt 10pt 20pt 20pt}, clip, width=\textwidth]{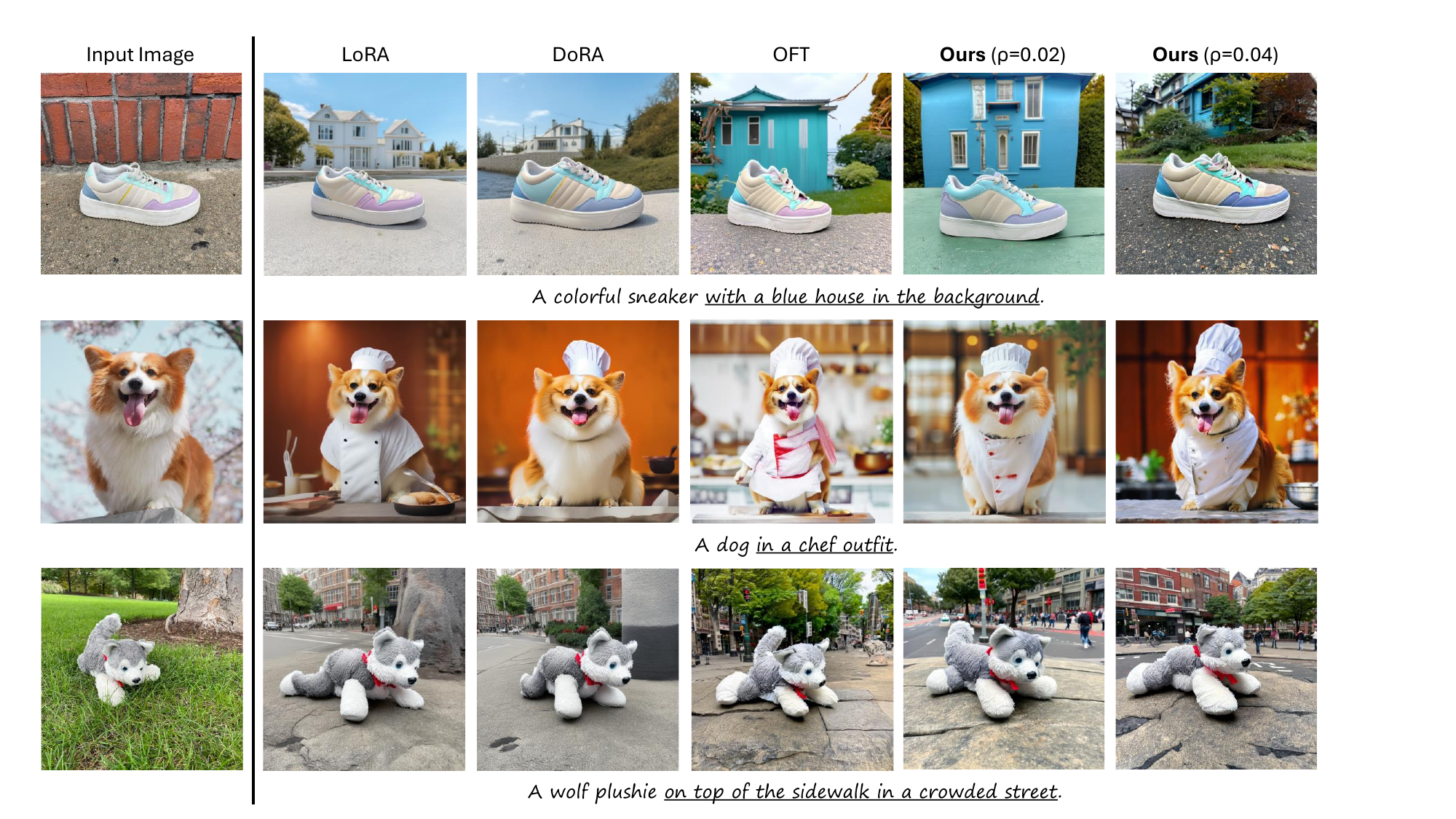}

 \end{subfigure}
 
  \caption{Personalization generated results comparison}
  \label{fig:personalization_3}
\end{figure*}

\begin{figure*}[t]
  \centering
 \begin{subfigure}[b]{1\textwidth}
     \centering
     \includegraphics[trim={0pt 10pt 20pt 20pt}, clip, width=\textwidth]{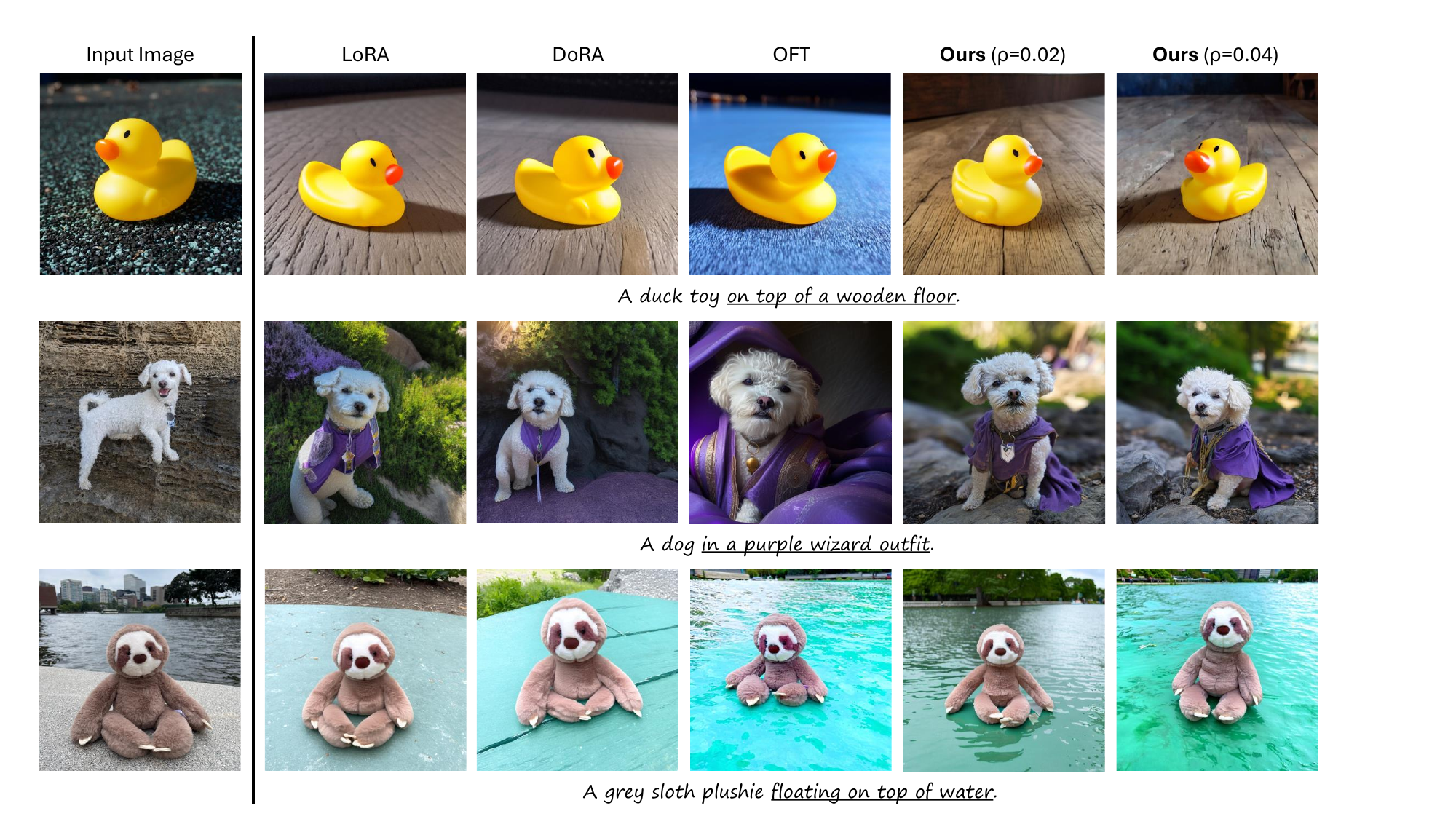}

 \end{subfigure}
 
  \caption{Personalization generated results comparison}
  \label{fig:personalization_4}
\end{figure*}

\begin{figure*}[t]
  \centering
 \begin{subfigure}[b]{1\textwidth}
     \centering
     \includegraphics[trim={0pt 10pt 20pt 20pt}, clip, width=\textwidth]{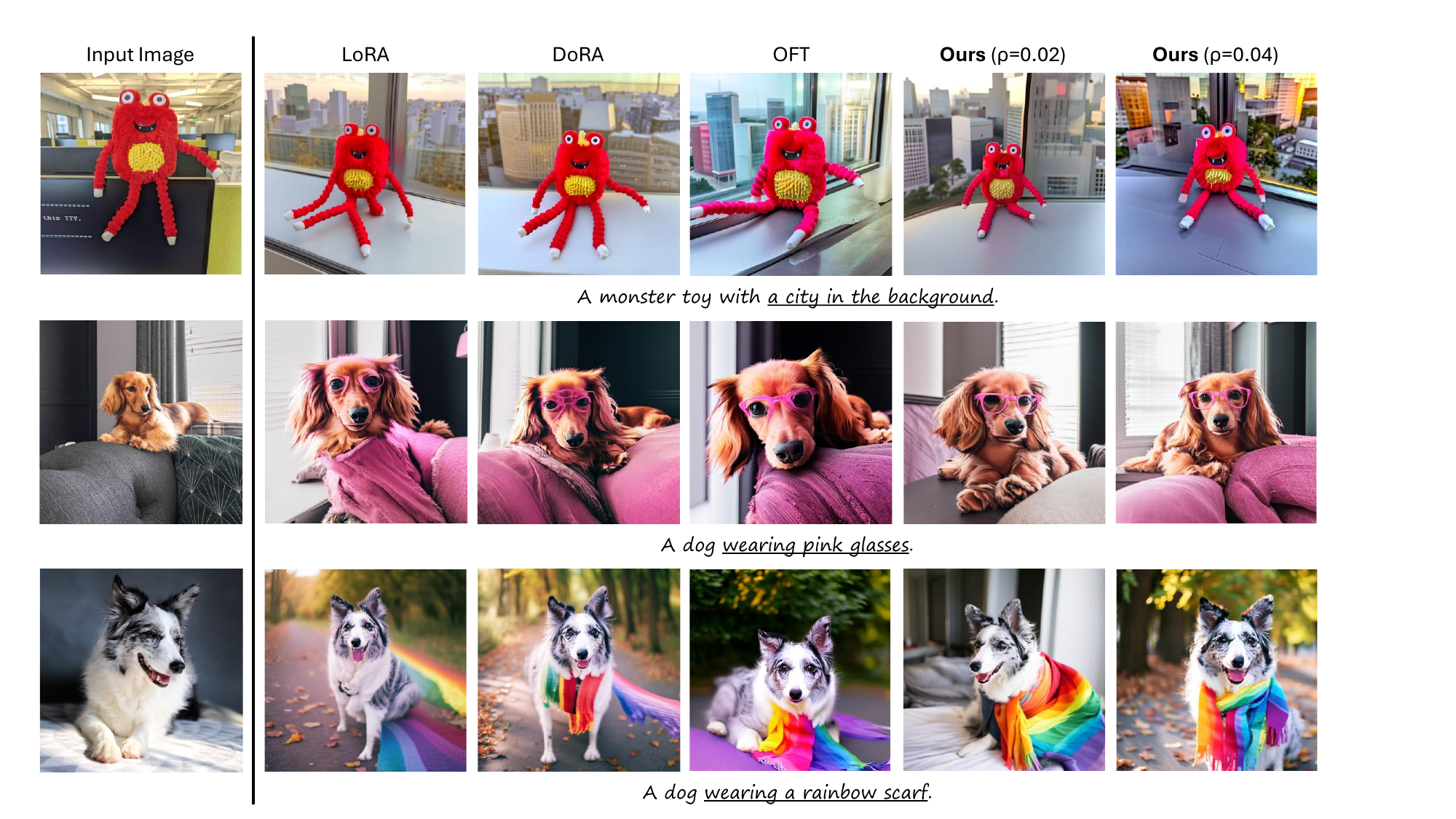}

 \end{subfigure}
 
  \caption{Personalization generated results comparison}
  \label{fig:personalization_5}
\end{figure*}

\begin{figure*}[t]
  \centering
 \begin{subfigure}[b]{1\textwidth}
     \centering
     \includegraphics[trim={0pt 10pt 20pt 20pt}, clip, width=\textwidth]{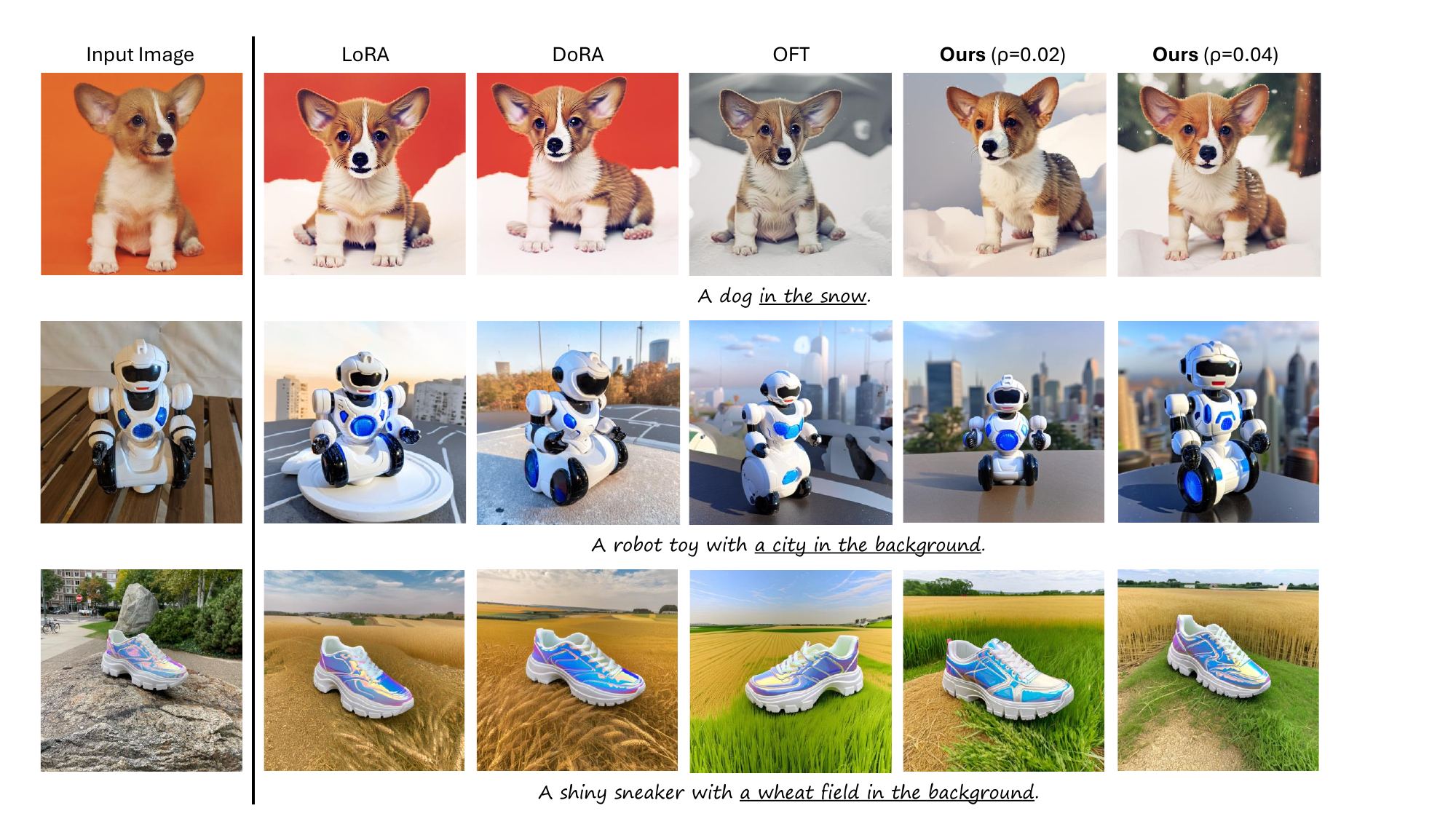}

 \end{subfigure}
 
  \caption{Personalization generated results comparison}
  \label{fig:personalization_6}
\end{figure*}

As shown in Figures~\ref{fig:personalization_1}–\ref{fig:personalization_6}, our approach produces outputs that not only align more accurately with the text prompts but also preserve the fine details of each learned concept more effectively than the baselines.

\end{document}